\documentclass[nohyperref]{article}

\usepackage{microtype}
\usepackage{graphicx}
\usepackage{booktabs} 

\usepackage{times}  
\usepackage{helvet}  
\usepackage{courier}  
\usepackage[hyphens]{url}  
\usepackage{graphicx} 
\usepackage{natbib}  
\usepackage{caption} 
\usepackage{algorithm}
\usepackage{algorithmic}
\usepackage{booktabs}
\usepackage{amsmath}
\usepackage{amssymb}
\usepackage{mathtools}
\usepackage{amsthm}
\usepackage{xcolor} 
\usepackage{bm}
\usepackage{longtable}
\usepackage{array}
\usepackage{multirow}
\usepackage{float}
\usepackage{enumitem}
\usepackage{diagbox}

\usepackage{hyperref}

\usepackage[accepted]{icml2022}

\usepackage{graphicx}
\usepackage{subcaption}
\usepackage{amsmath}
\usepackage{amssymb}
\usepackage{mathtools}
\usepackage{amsthm}
\usepackage{lipsum}
\usepackage{bbding}
\usepackage{colortbl}
\definecolor{mygreen}{HTML}{036400}
\definecolor{myred}{HTML}{cb0000}
\definecolor{mygray}{HTML}{999999}
\usepackage[capitalize,noabbrev]{cleveref}
\usepackage{makecell}
\usepackage{hyperref}
\usepackage{float}
\theoremstyle{plain}

\theoremstyle{definition}

\theoremstyle{remark}

\usepackage[textsize=tiny]{todonotes}
\usepackage{color}

\icmltitlerunning{SaFormer: A Conditional Sequence Modeling Approach to Offline Safe Reinforcement Learning}

\begin{document}

\twocolumn[
\icmltitle{SaFormer: A Conditional Sequence Modeling Approach to\\ Offline Safe Reinforcement Learning}



\icmlsetsymbol{equal}{*}

\begin{icmlauthorlist}
\icmlauthor{Qin Zhang}{equal,yyy}
\icmlauthor{Linrui Zhang}{equal,yyy}
\icmlauthor{Haoran Xu}{comp2}
\icmlauthor{Li Shen}{comp}
\icmlauthor{Bowen Wang}{yyy}
\icmlauthor{Yongzhe Chang}{yyy}
\icmlauthor{Xueqian Wang}{yyy}
\icmlauthor{Bo Yuan}{xxx}
\icmlauthor{Dacheng Tao}{comp}
\end{icmlauthorlist}

\icmlaffiliation{yyy}{Tsinghua University}
\icmlaffiliation{comp}{JD Explore Academy}
\icmlaffiliation{comp2}{JD Technology}
\icmlaffiliation{xxx}{Qianyuan Institute of Sciences}

\icmlcorrespondingauthor{Li Shen}{mathshenli@gmail.com}

\icmlkeywords{Machine Learning, ICML}

\vskip 0.3in
]



\printAffiliationsAndNotice{\icmlEqualContribution} 

\begin{abstract}
Offline safe RL is of great practical relevance for deploying agents in real-world applications. However, acquiring constraint-satisfying policies from the fixed dataset is non-trivial for conventional approaches. Even worse, the learned constraints are stationary and may become invalid when the online safety requirement changes. In this paper, we present a novel offline safe RL approach referred to as SaFormer, which tackles the above issues via conditional sequence modeling. In contrast to existing sequence models, we propose cost-related tokens to restrict the action space and a posterior safety verification to enforce the constraint explicitly. Specifically, SaFormer performs a two-stage auto-regression conditioned by the maximum remaining cost to generate feasible candidates. It then filters out unsafe attempts and executes the optimal action with the highest expected return. Extensive experiments demonstrate the efficacy of SaFormer featuring (1) competitive returns with tightened constraint satisfaction; (2) adaptability to the in-range cost values of the offline data without retraining; (3) generalizability for constraints beyond the current dataset.
\end{abstract}

\section{Introduction}
Reinforcement learning (RL)-based agent has achieved impressive performance in simulations~\citep{mnih2015human,silver2017mastering,vinyals2019grandmaster} but may adopt illegal actions and incur potential damage to the surroundings when it comes to the real world. In most safety-critical applications, agents are expected to follow certain constraints, such as maximum energy flow in smart grids~\citep{koutsopoulos2011control} and traffic regulations in autonomous driving~\citep{sallab2017deep}. Consequently, safe RL (a.k.a constrained RL) has gained extensive traction in recent years. However, most of the solutions~\citep{achiam2017constrained,yang2020projection,liu2020ipo} satisfy the constraints via  trial and error. Although the final policy may adhere to the safety requirement, the training process itself is still not risk-free. 

In a risky or costly environment, it is more applicable to learn constraint-satisfying policies from existing offline data~\citep{levine2020offline} instead of online interactions~\cite{ray2019benchmarking}. Nevertheless, addressing the intersection of safe RL and offline RL is non-trivial. First, the dataset may contain mixed, unsafe, or even conflicting demonstrations, making safety identification cumbersome. Second, the estimation for a long-term cost return is notoriously insufficient, which inevitably leads to infeasible or sub-optimal policies. Third, the intertwined optimization is extremely unstable when dual variables are introduced to handle the constraints.

In this paper, we tackle the above issues by leveraging sequence modeling, which is of independent interest in recent offline RL literature~\citep{chen2021decision,janner2021offline}. This paradigm sidesteps Bellman backups and shows the promise for a reliable cost estimation for safety-critical tasks. However, applying the current art to offline safe RL is still intractable. Take Decision Transformer (DT)~\citep{chen2021decision} as an example. It makes sense to increase the reward-to-go (RTG) manually for higher rewards, but that might be detrimental when considering the constraints: The agent demands more information to navigate the trade-off between performance and safety in the constrained action space.

We present a novel architecture referred to as SaFormer, which, to the best of our knowledge, is the first sequence modeling approach to offline safe RL. First of all, we propose two cost-related tokens, namely the cost limit token and the cost-to-go (CTG) token. The former is the human-specified constraint threshold defined at the beginning of an episode; the latter indicates the residual cost that the agent can still afford at the current time step. SaFormer exploits these two tokens and the past sequence to determine the target RTGs and then generates the feasible actions at the next stage. Another attractive benefit is that we can treat the specific threshold as a flexible prompt for SaFormer during the execution instead of retraining the policy toward different online safety requirements. Furthermore, SaFormer does not completely rely on the unidirectional sequence generation to achieve zero constraint violation. Instead, it samples a batch of candidates from the stochastic distribution, filters out unsafe attempts, and eventually executes the feasible action with the highest expected return. 

In the experiments, we follow a challenging but more practical pattern: The offline data acquisition is cost-agnostic, based on arbitrary behaviour policies instead of a mixture of pre-trained safe and unsafe policies. We customize safety constraints hind-sightly and relabel the dataset with specific cost criteria, which facilitates the reuse of offline data.

We conduct experiments on D4RL benchmarks~\citep{fu2020d4rl} subjecting to constraints on the cumulative torque. Empirical results confirm the strengths of our approach as: 
\vspace{-0.2cm}
\begin{enumerate}[leftmargin=1.25em]
\item SaFormer outperforms state-of-the-art baselines for competitive reward improvements and strict constraint satisfactions in both discounted and undiscounted settings.
\item SaFormer treats the constraint as contextual tokens and is adaptive to different cost limits in the range of existing offline data without retraining.
\item SaFormer supports online fine-tuning to satisfy the unexplored constraints beyond the current dataset.
\end{enumerate}

\section{Related Work}


\subsection{Offline Safe Reinforcement Learning}
Offline safe RL is an intersection of safe RL~\citep{altman1999constrained} and offline RL~\citep{levine2020offline}. Conventional methods~\cite{tessler2018reward,wang2020critic} are less effective when dealing with constrained optimization and conservative learning simultaneously. \citet{le2019batch} are the first to study the problem of batch policy learning under constraints. They simply utilize the off-policy evaluation for safety constraints, which limits their approach to the discrete action space. Based on the above work, \citet{polosky2022constrained} propose a projection algorithm that corrects the rewards-optimal policy back to the cost-feasible set through the Fenchel duality. Their analysis explicitly accounts for the distributional shift and offers non-asymptotic confidence bounds on the cost return. COptiDICE~\citep{lee2021coptidice} instead directly estimates the stationary distribution corrections of the optimal policy and therefore yields a single solvable optimization objective. Orthogonal to existing approaches, \citet{xu2022constraints} trains an independent safety critic, which penalizes unsafe actions and disables policy updates on them. Their proposed CPQ algorithm achieves impressive performance on a set of constrained locomotion tasks in continuous spaces. Compared with the prior work, our sequence modeling approach sidesteps cumbersome conservative Bellman backups and can be more flexible in dealing with varying constraint thresholds in the execution.

\subsection{RL via Sequence Modeling}
Sequence modeling has been widely adopted in natural language processing~\citep{radford2018improving} and computer vision~\citep{dosovitskiy2020image}. Recently, casting offline RL as a sequence generation problem shows the promise as well \citep{hu2022transforming}. Trajectory Transformer (TT)~\citep{janner2021offline} discretizes the state-action-reward sequence and  maximizes the sampling probability via beam search. Moreover, the transformer-based architecture features more reliable long-horizon prediction ability than single-step models, which also enables a fake trajectory bootstrapping for offline data augmentation~\citep{wang2022bootstrapped}. As a concurrent work, Decision Transformer (DT)~\citep{chen2021decision} utilizes the future return to condition the sequence generation. Their model-free method reduces the computational burden and generally outperforms TT in terms of cumulative rewards. \citet{xu2022prompting} pre-train DT on a large-scale dataset and realizes few-shot generalization by providing demonstrations. 
\citet{pmlr-v162-zheng22c} propose a stochastic DT to overcome the gap between offline training and online fine-tuning. 
To our best knowledge, there is no existing work that relates sequence modeling to offline safe RL.
\section{Preliminaries}
In this section, we first provide our problem formulation under the constrained MDP framework. We then briefly revisit the state of the art approach in sequential decision-making and discuss its applicability to offline safe RL. 

\subsection{Problem Formulation}
In this paper, we are interested in acquiring the constraint-satisfying policy from a stationary offline dataset. The problem formulation follows the constrained Markov Decision Process (CMDP)~\citep{altman1999constrained} denoted by a tuple $(\mathcal{S},\mathcal{A},\mathcal{P},\rho_0,\mathcal{R},\mathcal{C })$.
Here, ${\mathcal{S}}$ and ${\mathcal{A}}$ represent the state space and the action space, respectively.
${\mathcal{P}(s'|s,a)}: {\mathcal{S}} \times \mathcal{A} \times \mathcal{S} \mapsto [0,1]$ accounts for the state transition probability after applying action $a$ to the environment. 
$\rho_0:\mathcal{S} \mapsto [0,1]$ is the initial state distribution. 
$\mathcal{R} : \mathcal{S} \times \mathcal{A} \mapsto \mathbb{R}$ and $\mathcal{C} : \mathcal{S} \times \mathcal{A} \mapsto \mathbb{R}^+$ denote the reward and cost functions. 
A policy $\pi : S \mapsto P(A)$ maps the current state to a distribution over the action space. 
The discounted cumulative return of policy $\pi$ can be calculated by $J_R(\pi) = \mathop{\mathbb{E}}_{\tau}\big [ \sum^\infty_{t=0}\gamma^t r(s_t ,a_t)\big ]$, 
where $\tau = (s_0,a_0,s_1,...)$ is determined by $s_0 \sim \rho_0(\cdot), a_t \sim \pi(\cdot | s_t ), s_{t+1} \sim \mathcal{P}(\cdot | s_t,a_t)$. 
Similarly, we define the discounted cost return as $J_C(\pi) = \mathop{\mathbb{E}}_{\tau}\big [ \sum^\infty_{t=0}\gamma_c^t r(s_t ,a_t)\big ]$. 
 Formally, the constrained RL problem is given by
\begin{equation}
\mathop{\max}_{\pi} J_R(\pi)\quad  \mathrm{s.t.}\ \  J_C(\pi) \leq d.
\end{equation}
The agent is excepted to limit the cost return down to the threshold $d$; otherwise, the policy is deemed infeasible.

Note that, in our offline setting, the agent is not accessible to the environment. We can only tackle the above constrained problem using a fixed dataset $\mathcal{D} = \{\tau_i\}_{i=1}^N$, collected from a range of behaviour policies $\pi_\beta$. 

We assume that our offline safe RL tasks feature the following properties. First, the reward and cost returns are at least weakly correlated, which is essential to restrict the expected return via cost tokens. This assumption is mild since risk usually comes with profit. Second, the offline data contains trajectories with heterogeneous cost returns. This assumption is a necessary condition for our safety identification and action proposal. Instead, our method may not be distinguished from imitation learning (IL) when learning from completely homogeneous demonstrations, for example, all the trajectories are cost-signal-free and the expected constraint threshold is equal to $0$ as well.

\subsection{Revisit Decision Transformer in Offline Safe RL}
Decision Transformer (DT)~\citep{chen2021decision} brings a novel sequence modeling paradigm to offline RL. It employs 3 types of tokens, namely $\hat{R}_t,s_t,a_t$, and represent a trajectory $\tau$ in order as $\langle \hat{R}_0, s_0, a_0, \hat{R}_1, s_1, a_1, . . . , \hat{R}_T, s_T, a_T \rangle$. 
Here, the return-to-go (RTG) token $\hat{R}_t = \sum^T_{t'=t} r_{t'}$ denotes the sum of future rewards from the current step $t$, which is the pivotal prompt for the auto-regressive sequence generation to achieve desired performance. 
At each timestep $t$, DT is fed with $3K$ tokens from the last $K$ timesteps and then predicts the following deterministic action $ a_t = \pi_\mathrm{DT}(\{\hat{R}_{i}, s_{i}, a_i\}^{t-1}_{t-K} \cup \{\hat{R}_{t-k}, s_{t-k}\})$ via a casually masked Transformer. Here, $\{\hat{R}_{i}, s_{i}, a_i\}^{t-1}_{t-K}$ is as a shorthand for the subsequence $\langle \hat{R}_{\max\{0.t-K\}},s_{\max\{0.t-K\}},$ $ a_{\max\{0.t-K\}},...,\hat{R}_{t-1}, s_{t-1}, a_{t-1}\rangle.$
The policy is learned by minimizing the $\ell_2$ objective as follows:
\begin{equation}
\label{eq:dt}
\resizebox{.91\linewidth}{!}{$
     \mathbb{E}_\mathcal{D} \sum_{k=0}^K \big( a_{t-k} - \pi_\mathrm{DT}(\{\hat{R}_{i}, s_{i}, a_i\}^{t-1-k}_{t-K} \cup \{\hat{R}_{t-k}, s_{t-k}\}) \big)^2. $}
\end{equation}
In the phase of policy evaluation, DT is specified with the desired episodic return $\hat{R}_0$ at the initial state $s_0$ and then generates $a_0$.
Once $a_0$ is performed, the agent observes the next state $s_1$ and the reward $r_0$, which gives $\hat{R}_1 = \hat{R}_0 - r_0$. These tokens are sufficient for the following sequence generation, and the loop repeats until the episode terminates.

\begin{figure}[ht]
\begin{center}
\centerline{\includegraphics[width=\columnwidth]{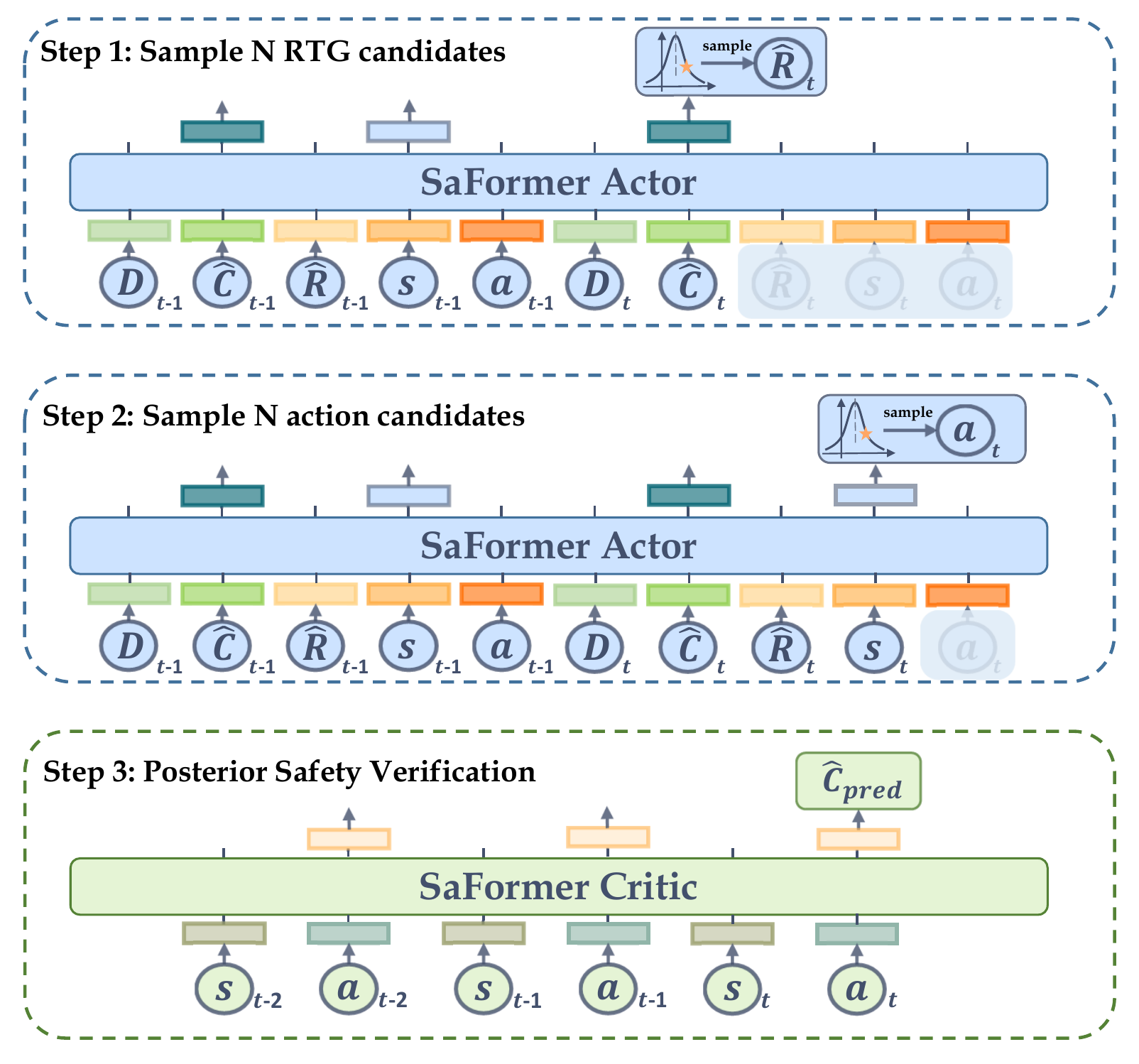}}
\caption{The schema of SaFormer. The actor (top \& middle) first generates the RTG distribution via cost-related tokens and then generates action candidates after sampling potential RTGs. The critic (bottom) predicts the long-term cost return of each candidate to filters out unsafe attempts and eventually selects the feasible action with the highest future reward designation.}
\label{framework}
\end{center}
\vskip -0.2in
\end{figure}

\citet{chen2021decision} show that the eventual return strongly correlates with the initial RTG which can even be extrapolated to values beyond the maximum in the dataset. Nevertheless, tweaking RTG manually in offline safe RL is not straightforward since the optimal RTG is heavily dependent on the constraints. Raising RTG aggressively may incur infeasible actions that conflict with certain safety constraints; while degrading RTG conservatively may lead to sub-optimal policies. In summary, DT has to be conditioned via cost-reliant RTGs and other informative tokens to navigate the trade-off between performance and safety.

\section{Methodology}
In this section, we present SaFormer, which addresses the above issues with the conditional sequence generation and the posterior safety verification, as illustrated in \cref{framework}.

\subsection{Feasible Action Proposal}

Considering the cost constraints in offline safe RL, we propose two  cost-related tokens, namely the cost-to-go (CTG) $\hat{C}_t = \sum^T_{t'=t} c_{t'}$ and the cost limit $D = d$. 

The CTG accounts for residual costs that the agent can still afford from now on, which is crucial to condition the action. 
Even though the update of CTG is computable during the execution since the initial value is equal to the constraint threshold at the beginning of an episode, we still expect that predicting CTGs as an auxiliary task in the training process  is beneficial for learning better representations. The motivation comes from the fact that it is equivalent to learning the cost function considering that $c_t = \hat{C}_{t+1} -  \hat{C}_{t}$.

The cost limit $D$ is the constraint threshold and held constant throughout the sequence. We keep the token $D$ instead of feeding CTG only since $D -\hat{C}_t $ indicates the cumulative cost  incurred before. It contains past behavioral information that better conditions future sequence generation for a more general partially observable Markov decision process.

Under the assumption that the cost limit token $D$, the CTG token $\hat{C}_t$, and the past subsequence $\{D,\hat{C}_i,\hat{R}_{i}, s_{i}, a_i\}^{t-1}_{t-K}$ are sufficient to determine the current RTG target, SaFormer generates RTG distribution at timestep $t$ as follows:
\begin{equation}
\small
    \hat{R}_t \sim \mathcal{N}(\mu_\theta,\Sigma_\theta | \{D,\hat{C}_i,\hat{R}_{i}, s_{i}, a_i\}^{t-1}_{t-K} \cup \{D,\hat{C}_t\} ).
\end{equation}

\begin{algorithm}[tb]
\small
   \caption{SaFormer Policy Execution}
   \label{alg:saformer1}
\begin{algorithmic}[1]
    \STATE {\bfseries Require:} Online $Env$, Actor $\pi_\theta$, Critic $\zeta_\phi$, cost limit $d$.
   \STATE Initialize $D = \hat{C}_0 = d$.
   \FOR{$t=0$ {\bfseries to} maximum episode horizon $T$}
   \STATE \texttt{\#Feasible Action Proposal}
   \STATE SaFormer actor $\pi_\theta$ generates the RTG distribution  $\mathcal{N}(\mu_\theta,\Sigma_\theta | \{D,\hat{C}_i,\hat{R}_{i}, s_{i}, a_i\}^{t-1}_{t-K} \cup \{D,\hat{C}_t\} )$.
   \STATE Sample $N$ RTG candidates $\hat{R}^n_t, n=1,2,...,N$.
   \STATE SaFormer actor $\pi_\theta$ generates $N$ action distributions $\mathcal{N}(\mu_\theta,\Sigma_\theta | \{D,\hat{C}_i,\hat{R}_{i}, s_{i}, a_i\}^{t-1}_{t-K} \cup\{D,\hat{C}_t,\hat{R}^n_t,s_t\} )$.
   \STATE Sample $N$ action candidates $a^n_t, n=1,2,...,N$.
   \STATE \texttt{\#Posterior Safety Verification}
   \STATE SaFormer critic $\zeta_\phi$ predicts the long-term cost-return of each pair $<\hat{R}^n_t, a^n_t>, n = 1,2,...,N$.
   \IF{$\exists n, \zeta_\phi(\{s_{i}, a^n_i\}^{t}_{t-K}) \leq \hat{C}_t$ }
   \STATE $a_t = \arg\max(\hat{R}^n_t|\zeta_\phi(\{s_{i}, a^n_i\}^{t}_{t-K}) \leq \hat{C}_t)$
   \ELSE
   \STATE Reject and re-sample $<\hat{R}^n_t, a^n_t>, n = 1,2,...,N$.
   \ENDIF
   \STATE Update $s_t,r_t,c_t = Env.step(a_t)$ and
   $\hat{C}_t = \hat{C}_t - c_t$.
   \ENDFOR
\end{algorithmic}
\end{algorithm}

After sampling the target from the RTG distribution, we then generate the action distribution next based on the past sequence plus the future reward expectation $\hat{R}_t$ and the visiting state $s_t$ as follows:
\begin{equation}
\small
    \hat{a}_t \sim \mathcal{N}(\mu'_{\theta},\Sigma'_{\theta} | \{D,\hat{C}_i,\hat{R}_{i}, s_{i}, a_i\}^{t-1}_{t-K} \cup \{D,\hat{C}_t,\hat{R}_t,s_t\} ).
\end{equation}

The above probabilistic mapping is more reasonable than a deterministic model since the agent can perform different actions to achieve the same cost return, which yields a wide range of feasible RTGs under a definite CTG.
Apart from that, it enables SaFormer to optimize CTGs, RTGs, and actions within a single auto-regressive model. Tweaking multi-objective $\ell_2$ loss function in \cref{eq:dt} is cumbersome considering that the scales of the above tokens vary significantly among different scenarios. Instead, we keep them aligned via a multivariate independent Gaussian distribution $P(\hat{C}_t,\hat{R}_t,a_t | \{D,\hat{C}_i,\hat{R}_{i}, s_{i}, a_i\}^{t-1}_{t-K})$ and minimize the negative log-likelihood (NLL) objective $J(\theta)$ as follows:
\begin{equation}
\label{eq:actor}
\resizebox{.91\linewidth}{!}{$
    \begin{aligned}
\!\!\!\! J(\theta)& \!=\!  -\frac{1 }{K\!+\!1} \mathbb{E}_\mathcal{D} \sum_{k=0}^K \big[ 
         \log \pi_\theta \big( \hat{C}_{t-k} |  
        \{D,\hat{C}_i,\hat{R}_{i}, s_{i}, a_i\}^{t-1-k}_{t-K} \cup D
        \big)\\
         & + \log \pi_\theta \big( \hat{R}_{t-k} |  
        \{D,\hat{C}_i,\hat{R}_{i}, s_{i}, a_i\}^{t-1-k}_{t-K} \cup \{D,\hat{C}_{t-k}\}
        \big)\\
        & + \log \pi_\theta \big( a_{t-k} | 
        \{D,\hat{C}_i,\hat{R}_{i}, s_{i}, a_i\}^{t-1-k}_{t-K} \cup \{D,\hat{C}_{t-k},\hat{R}_{t-k},s_t\}
        \big) \big].
    \end{aligned}
    $}
\end{equation}

Another attractive point of this design is that the stochastic action distribution naturally fits in with online exploration techniques~\citep{pmlr-v162-zheng22c} and has the potential to transfer to unexplored constraints beyond the offline dataset, which we will discuss in \cref{4.3}.

\subsection{Posterior Safety Verification}

The cost-conditioned sequence generation is effective in producing near-optimal solutions. Nevertheless, we argue that it is impractical to solely rely on naive forward-computing to satisfy the hard constraint. Instead, the unidirectional auto-regression in our approach only nominates the potential candidates, and an additional transformer-based critic $\zeta_\phi$ is proposed to enforce safety explicitly. Although the optimization of $\zeta_\phi$ is irrelevant with Bellman backups in the conventional actor-critic architecture~\cite{fujimoto2019off,wang2020critic}, we inherit the name ``critic'' since it evaluates the cost that the agent may incur in the future. The critic $\zeta_\phi$ is updated via causal sequence modeling, and the self-supervised objective is equal to the CTG in the dataset:
\begin{equation}
\label{eq:critic}
    \small
    \widetilde{J}(\phi) = 
 \frac{1}{K+1}\mathbb{E}_\mathcal{D} \sum_{k=0}^K \big( \hat{C}_{t-k} - \zeta_\phi(\{s_{i}, a_i\}^{t-k}_{t-K}) \big)^2. 
\end{equation}

\begin{algorithm}[tb]
\small
   \caption{SaFormer Offline Training}
   \label{alg:saformer2}
\begin{algorithmic}[1]
   \STATE {\bfseries Require:} Offline dataset $\mathcal{D}$, Actor $\pi_\theta$, Critic $\zeta_\phi$
   \REPEAT
  
   \FOR{$iter=0$ {\bfseries to} maximum iteration $M$}
    \STATE Sample a mini-batch $\{D,\hat{C}_i,\hat{R}_{i}, s_{i}, a_i\}^{t}_{t-K} \sim \mathcal{D}$.
   \STATE Optimize SaFormer actor $\pi_\theta$ via minimizing Eq~\eqref{eq:actor}.
   \STATE Optimize SaFormer critic $\zeta_\phi$ via minimizing Eq~\eqref{eq:critic2}.
   \ENDFOR
   \STATE Policy evaluation using~\cref{alg:saformer1}.
   \UNTIL{the offline training terminates.}
\end{algorithmic}
\vspace{-0.025cm}
\end{algorithm}

\begin{algorithm}[tb]
\small
   \caption{SaFormer Online Fine-tuing}
   \label{alg:saformer3}
\begin{algorithmic}[1]
   \STATE {\bfseries Require:}  Online  $Env$,  Offline dataset $\mathcal{D}$, Actor $\pi_\theta$, Critic $\zeta_\phi$, OOD cost limit $d$. Attenuation factor $\alpha \in (0,1)$.
     \REPEAT
    \STATE Set $d = \alpha d$ as the target cost limit.
  \STATE Rollout trajectory $\tau$ using~\cref{alg:saformer1}.
  \STATE Relabel $\tau$ as $\hat{R}_t = \sum^T_{t'=t} r_{t'}, \hat{C}_t = \sum^T_{t'=t} c_{t'}, D = \hat{C}_0$.
  \STATE Aggregate the offline data $\mathcal{D} = \mathcal{D} \cup \tau$.
   \STATE Fine-tune SaFormer $\pi_\theta$ and $\zeta_\theta$ using~\cref{alg:saformer2}.
   \UNTIL{the online fine-tuning terminates.}

\end{algorithmic}
\vspace{-0.025cm}
\end{algorithm}

In our implementation, we introduce the regularization terms into the objective function as:
\begin{equation}
\label{eq:critic2}
\resizebox{.91\linewidth}{!}{$
\begin{aligned}
    J(\phi)= \widetilde{J}(\phi) + \lambda\cdot\sum_{k=0}^{K-1}[ \zeta_\phi(\{s_{i}, a_i\}^{t-k}_{t-K}) - \zeta_\phi(\{s_{i}, a_i\}^{t-k-1}_{t-K})]^+.
\end{aligned}
$}
\end{equation}

The penalized objective leads to more reasonable representations since the cost signal $c$ is always non-negative, and the sequence of CTGs is monotonically decreasing.

Consequently, the critic $\zeta_\phi$ serves as the CTG estimator and conducts a posterior safety verification after the actor $\pi_\theta$ has proposed feasible action candidates. As shown in~\cref{framework}, the execution of SaFormer consists of the following steps: 
\begin{enumerate}
\vspace{-0.15cm}
\item SaFormer actor generates the RTG distribution from $\{D,\hat{C}_i,\hat{R}_{i}, s_{i}, a_i\}^{t-1}_{t-K} \cup \{D,\hat{C}_t\}$ and samples $N$ RTG candidates $\hat{R}^n_t, n=1,2,...,N$.
\item SaFormer actor consumes $N$ RTG candidates to generates $N$ corresponding action distributions from $\{D,\hat{C}_i,\hat{R}_{i}, s_{i}, a_i\}^{t-1}_{t-K} \cup \{D,\hat{C}_t,\hat{R}^n_t,s_t\}$. Then, it samples $N$ action candidates $a^n_t$ from the distributions.
\item SaFormer critic evaluates the long-term cost-return of each pair $\{\hat{R}^n_t, a^n_t\}, n = 1,2,...,N$ to filter out unsafe attempts $\zeta(\{s_{i}, a^n_i\}^{t}_{t-K}) > \hat{C}_t$, and executes the one with the highest $\hat{R}_t$ among the remaining actions.
\end{enumerate}
\vspace{-0.2cm}
Details of online execution and offline training for SaFormer are summarized in Algorithms \ref{alg:saformer1} and \ref{alg:saformer2}, respectively.

\subsection{Online Fine-tuning for OOD Constraints}
\label{4.3}
\begin{figure}
\begin{center}
\centerline{\includegraphics[width=0.95\columnwidth,trim=0 0 0 0,clip]{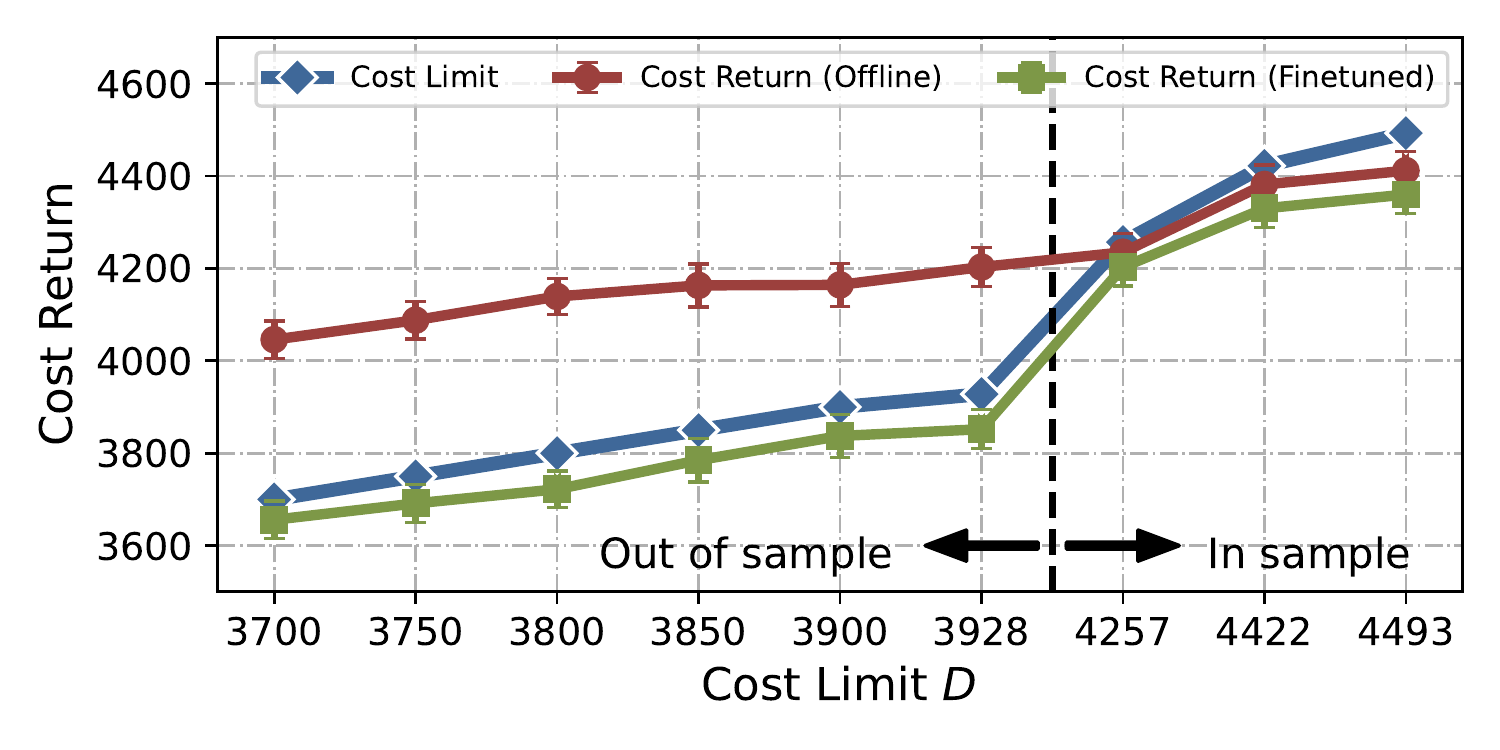}}
\vspace{-0.35cm}
\caption{Cost returns of SaFormer after offline pre-training (red) and online fine-tuning (green) with respect to a range of expected cost limits (blue) of the HalfCheetah\_medium\_replay dataset.}
\label{onlineexp}
\end{center}
\vspace{-0.9cm}
\end{figure}

SaFormer treats different cost limits as contextual tokens; thus, it is applicable to the in-range constraint values of the existing dataset without retraining. It also shows the promise for constraint satisfaction beyond the offline data distribution considering the impressive generalization ability of Transformer architecture~\citep{vaswani2017attention}. \cref{onlineexp} shows that the final cost return strongly correlates with the specified cost limit and monotonically decreases when the threshold is tightened. Nevertheless, pure offline training might be insufficient for out-of-distribution (OOD) constraints when the collected data is limited. 

We further propose an online fine-tuning algorithm inspired by \citet{pmlr-v162-zheng22c}. Different from their purpose for boosting reward, we aim to fine-tune SaFormer for the tightened OOD constraints.
Specifically, we reformulate the naive NLL objective in \cref{eq:actor} under the MaxEnt RL framework~\citep{haarnoja2018soft} to benefit the exploration, which boils down to the following  constrained optimization:
\begin{equation}
    \mathop{\min}_{\theta} J(\theta)\quad \mathrm{s.t.} \ \ H(\theta) \geq \beta.
\end{equation}
Here, $ H(\theta)$ denotes the Shannon entropy of the action distribution of the SaFormer actor, which is defined as: 
\begin{equation}
\resizebox{.91\linewidth}{!}{$
    H(\theta) =  \frac{1 }{K+1} \mathbb{E}_\mathcal{D} \sum_{k=0}^K H\big[\pi_\theta\big( a_{t-k} | 
        \{D,\hat{C}_i,\hat{R}_{i}, s_{i}, a_i\}^{t-k}_{t-K}
        \big)\big].
        $}
\end{equation}

Notably, we employ an attenuation factor $\alpha$ to decrease the cost limit every time SaFormer interacts with the environment. It enables the acquisition of heterogeneous samples to speed up the training, especially when the expected constraint is difficult to satisfy with the initial pre-trained SaFormer. Nevertheless, we use the actual cost return as the value of token $D$ when we relabel the newly collected trajectory. Furthermore, we simply aggregate the dataset with new samples instead of discarding the earliest trajectory~\cite{pmlr-v162-zheng22c} in order to maintain the previously learned constraints. Details of online fine-tuning for SaFormer are summarized in \cref{alg:saformer3}.

\section{Experiments}
In this section, we first introduce our experiment setup following a cost-agnostic data acquisition. We then present empirical results to verify the efficacy of SaFormer in terms of (1) competitive return with tightened constraint satisfaction; (2) robustness and flexibility toward different in-range constraint thresholds of the offline data; (3) online fine-tuning for satisfying the unexplored constraints. At last, we conduct the ablation study to better support our approach.

\begin{figure*}[ht]
\begin{center}
\centerline{\includegraphics[width=2\columnwidth,trim=0 0 0 0,clip]{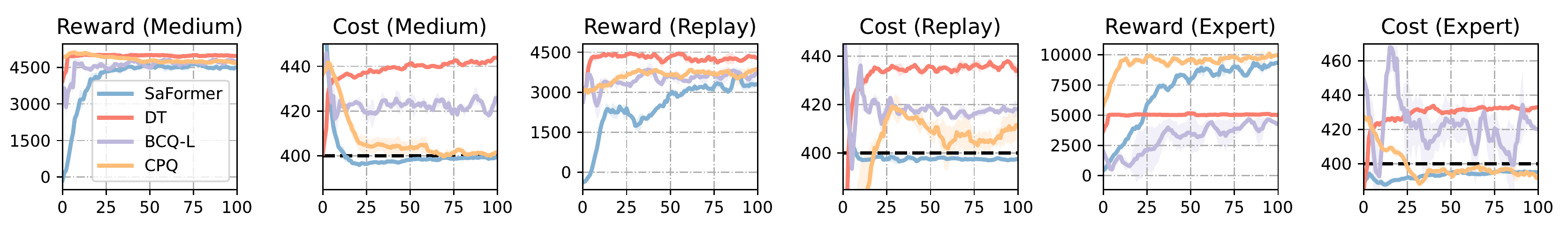}}
\vspace{-0.25cm}
\caption{Online evaluation curves on HalfCheetah datasets under the discounted setting ($\gamma_c = 0.99$). The x-axis denotes the number of offline training epochs ($5\times10^3$ iterations per epoch). The y-axis denotes the reward or cost return. The dashed line denotes the cost limit.}
\label{exp1}
\vskip 0.2in
\centerline{\includegraphics[width=2\columnwidth,trim=0 0 0 0,clip]{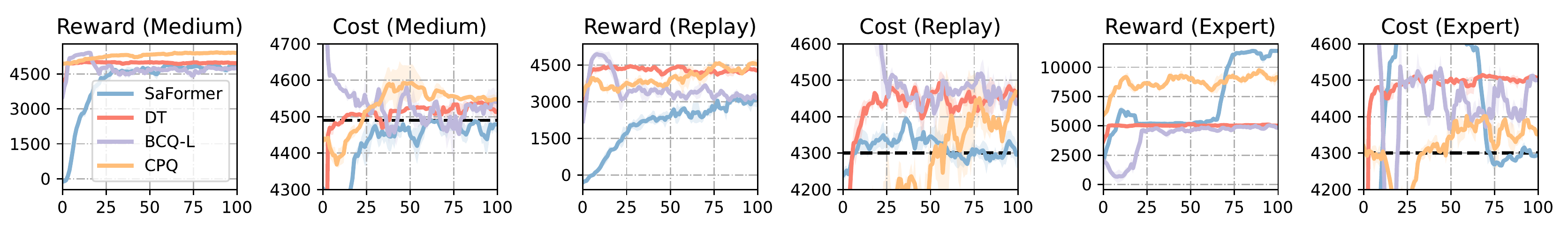}}
\vspace{-0.25cm}
\caption{Online evaluation curves on HalfCheetah datasets under the undiscounted setting ($\gamma_c = 1$). The x-axis denotes the number of offline training epochs ($5\times10^3$ iterations per epoch). The y-axis denotes the reward or cost return. The dashed line denotes the cost limit.}
\label{exp2}
\end{center}
\vskip -0.15in
\end{figure*}

\begin{table*}[t]
\begin{center}
\begin{small}
\begin{sc}
\renewcommand\arraystretch{1.15}
\resizebox{1\textwidth}{20.5mm}{
\begin{tabular}{|c|c|cccc|cccc|}
\hline
  \multicolumn{2}{|c|}{\multirow{2}*{Dataset}} &\multicolumn{4}{c|}{$\gamma_c=0.99$} &  \multicolumn{4}{c|}{$\gamma_c=1$} \\
  \cline{3-10}
  \multicolumn{2}{|c|}{~} & SaFormer & BCQ-L & CPQ & DT &  SaFormer & BCQ-L & CPQ & DT \\
\hline

\multirow{3}*{\makecell[c]{halfcheetah \\ medium}} 
&  reward    & $4519.21\pm 52.75$& $4734.96\pm 98.13$& \bm{$4727.55\pm 36.41$} &$4986.64 \pm 5.56$& \bm{$ 4661.20\pm 52.46$ }& $4729.00\pm 59.52$& $5411.98\pm 55.20$& $4986.64 \pm 5.56$\\
~ &  cost        &  \cellcolor{mygreen!20}$398.86\pm 0.62$& \cellcolor{myred!20}$421.68\pm 3.36$& \cellcolor{mygreen!20}$400.83\pm 0.79$&  $441.89\pm 0.21$ &\cellcolor{mygreen!20} $4487.2\pm 6.05$&\cellcolor{myred!20} $4529.94\pm 34.25$&\cellcolor{myred!20} $4548.34\pm 11.07$&  $4531.89\pm 1.45$ \\
~ &  limit  & $400$ & $400$ & $400$ & -- & $4490$ & $4490$ & $4490$ & -- \\

\hline

\multirow{3}*{\makecell[c]{halfcheetah \\ medium\_replay}} 
& reward    & \bm{$3321.55\pm 101.17$}& $3603.96\pm 144.22$& $3731.19\pm 163.03$& $4270.07 \pm 16.11$ &\bm{ $3019.17\pm 234.09$}& $3191.36\pm 101.62$& $4317.85\pm 192.30$& $4270.07 \pm 16.11$\\
~ &  cost        & \cellcolor{mygreen!20}$397.41\pm 0.43$& \cellcolor{myred!20}$417.86\pm1.98$& \cellcolor{myred!20}$409.59\pm 5.93$&  $436.16\pm 1.87$ & \cellcolor{mygreen!20}$4289.54\pm 6.31$&\cellcolor{myred!20} $4443.87\pm 49.28$& \cellcolor{myred!20}$4408.20\pm  51.61$&  $4448.84 \pm  6.8$ \\
~ &  limit  & $400$ & $400$ & $400$ & --& $4300$ & $4300$ & $4300$ & -- \\

\hline
\multirow{3}*{\makecell[c]{halfcheetah \\ medium\_expert}} 
& reward    & $9124.51\pm 78.08$& $4642.31\pm 234.34$& \bm{$9870.24\pm 169.63$}& $5053.89\pm 30.17$ &\bm{ $11000.37\pm 141.93$ }& $4902.81\pm 131.39$& $9221.09\pm 229.64$& $5053.89 \pm  30.17$\\
~ & cost        &\cellcolor{mygreen!20}$395.17\pm 0.79$& \cellcolor{myred!20}$425.13\pm 9.84$& \cellcolor{mygreen!20}$393.91\pm 1.57$&  $429.75\pm 0.37$ &  \cellcolor{mygreen!20}$4296.87\pm 9.92$&\cellcolor{myred!20} $4468.96\pm 50.47$& \cellcolor{myred!20}$4369.97\pm 18.63$&   $4478.89\pm 3.52$ \\
~ &  limit  & $400$ & $400$ & $400$ & -- & $4300$ & $4300$& $4300$ & --\\

\hline

\end{tabular}
}
\caption{Baseline comparison results on HalfCheetah datasets. The costs marked in \colorbox{mygreen!20}{\textcolor{mygreen!20}{xxx}} satisfy the constraint; the costs marked in \colorbox{myred!20}{\textcolor{myred!20}{xxx}} violate the constraint and the corresponding policies are deemed infeasible. We use the same notation in the rest part of the paper.} 
\label{tab:baselinecompare}
\end{sc}
\end{small}
\end{center}
\vspace{-0.25cm}
\end{table*}

\begin{table*}[t]
\begin{center}
\begin{small}
\begin{sc}
\renewcommand\arraystretch{1.15}
\resizebox{1\textwidth}{21.5mm}{
\begin{tabular}{|c|c|cccc|c|}
\hline
 \multicolumn{2}{|c|}{\small{Dataset}}  & \small{10\%Percentile} & \small{20\%Percentile} & \small{30\%Percentile}  & \small{50\%Percentile}  & DT(Reference)   \\
\hline

\multirow{3}*{\makecell[c]{halfcheetah \\ medium}} 
& reward    & $4661.2\pm 52.46$& $4713.52\pm 26.83$& $4718.44\pm 22.09$& $4724.89\pm 27.18$  & $4986.64\pm 5.56$ \\
~ & cost    & \cellcolor{mygreen!20}$4487.2\pm 6.05$& $ \cellcolor{mygreen!20}4498.22\pm 2.57$&  \cellcolor{mygreen!20}$4500.52\pm 1.65$& $ \cellcolor{mygreen!20}4508.83\pm 2.79$  & $4531.89\pm 1.45$ \\
~ &  limit  & $4490$ & $4503$ & $4511$ & $4526$ & --\\

\hline
\multirow{3}*{\makecell[c]{halfcheetah \\ medium\_replay}} 
& reward    & $2202.69\pm 148.39$& $2208.41\pm 90.14$ & $3363.11\pm 165.89$& $3712.08\pm 104.74$   & $4270.07 \pm 16.11$ \\
~ & cost    & \cellcolor{myred!20}$4203.0\pm 17.46$& \cellcolor{mygreen!20}$4234.15\pm 13.36$ &\cellcolor{mygreen!20} $4381.42\pm 16.94$& \cellcolor{mygreen!20}$4411.84\pm 9.29$ & $4448.84 \pm 6.8$\\
~ & limit   & $3928$ & $4257$ & $4422$ & $4493$ & --\\

\hline

\multirow{3}*{\makecell[c]{halfcheetah \\ medium\_expert}} 
& reward    & $9016.82\pm 124.45$& $9121.95\pm 397.5$& $9161.5\pm 385.65$& $9296.29\pm 249.52$ & $5053.89 \pm 30.17$\\
~ & cost    & \cellcolor{mygreen!20}$4181.25\pm 16.63$& $\cellcolor{mygreen!20}4203.95\pm 2.36$& $\cellcolor{mygreen!20}4211.91\pm 9.08$& $\cellcolor{mygreen!20}4229.4\pm 17.35$ &   $4478.89 \pm 3.52$ \\
~ & limit   & $4201$& $4215$ & $4224$ & $4266$ &  -- \\

\hline
\end{tabular}
}
\caption{Experiments on constraint adaptation. We train SaFormer on certain dataset and exploit different cost limit token in the execution. A full version of the results on all datasets can be found in \cref{appendixT2} \cref{tab:zero-shot}.}
\label{tab:zero-shot-main}
\end{sc}
\end{small}
\end{center}
\vspace{-0.25cm}
\end{table*}

\begin{table*}[t]
\begin{center}
\begin{small}
\begin{sc}
\renewcommand\arraystretch{1.15}
\resizebox{1\textwidth}{8.75mm}{
\begin{tabular}{|c|ccccc|cccc|}
\hline
 \multirow{2}*{Cost Limit $D$}  &\multicolumn{5}{c|}{Out of Sample} & \multicolumn{4}{c|}{In sample}  \\
  \cline{2-10}
  ~ & $3700$ & $3750$ & $3800$  & $3850$ & $3900$ & $3928$ & $4257$ & $4422$ & $4493$ \\
\hline

Avg Cost (offline) &  $4045.92$& $4088.0$& $4139.33$& $4163.35$& $4164.42$&  $4203.0$  & $4234.15$& $4381.42$&$4411.84$ \\

Avg Cost (finetuned) &$3655.93$& $3691.42$ & $3722.03$ &$3784.85$& $3837.36$&$3852.83$ & $4202.85$ &  $4330.46$ &  $4360.21$ \\
\hline
\end{tabular}
}
\caption{Experiments on constraint generalizability. We report the mean cost return of SaFormer after offline pre-trained on the HalfCheetah\_medium\_replay dataset and online fine-tuned with respect to a range of expected cost limits.}
\label{tab:online-main}
\end{sc}
\end{small}
\end{center}
\vspace{-0.25cm}
\end{table*}

\subsection{Experiment Setup}

\paragraph{Dataset}
We leverage the D4RL dataset~\cite{fu2020d4rl} over three Mujoco tasks, namely
\texttt{Hopper},  \texttt{Walker2d}, and \texttt{HalfCheetah}. Note that the behaviour policies are arbitrary and cost-agnostic when collecting the samples; the existing offline data is relabeled hind-sightly according to certain cost criteria. This paradigm is of great practical relevance in data reuse and can be more flexible for varying online safety requirements compared with the prior work~\cite{xu2022constraints,polosky2022constrained}. In previous study, the datasets are constructed by two types of behaviour policies: the unsafe one is trained with general RL methods; the safe one is trained via constrained RL adhering to a pre-defined constraint. Such the setting has several limitations: First, the cost distribution will center on two widely separated peaks, which is relatively easy to identify and thus can be simply solved by behavior cloning. Second, the agent can only learn a stationary constraint determined by the safe behavior policy. Most significantly, there would be a chicken-and-egg problem in practice for obtaining the constraint-satisfying policy in advance to construct the offline dataset.

\vspace{-0.2cm}
\paragraph{Task} Our safety consideration is to prolong the lifespan of motors. Therefore, we specify the cost as $c_t = \sum^{M}_{i=1} |a^i_t|$ standing for the total torque applied to the $M$ joints, and then we limit the cumulative energy consumption by enforcing the episodic constraint $C_\pi=\sum_{t=0}^T \gamma^t c_t \leq d$. We conduct experiments with $\gamma_c = 0.99$ and $\gamma_c = 1$, respectively. The discounted setting is widely adopted in conventional TD-learning algorithms and also fits in with SaFomer if the cost is relabeled as $c_t = \gamma_c^t c_t$; the undiscounted setting is more challenging but practical since the torque applied in the first and last steps are equally weighed for the lifespan of motors. 
As for the assignment of $d$, we sort the trajectory-wise cost return $C_{\pi_\beta}$ on each dataset and uniformly specify the $10\%,20\%,30\%,50\%$ percentiles as different thresholds. Readers can refer to \cref{appendixA} for the distributions of $C_{\pi_\beta}$ and calculations of $d$ regarding all the nine datasets.

\paragraph{Baselines} We compare our proposed SaFormer with the following three kinds of representative baselines:
\begin{itemize}[leftmargin=1.5em]
\vspace{-0.2cm}
\item DT~\citep{chen2021decision}: Decision Transformer with manually assigned RTGs is regarded as the unconstrained counterpart. We set a constant initial RTG = $3600,5000,6000$ for each dataset of \texttt{Hopper},  \texttt{Walker2d}, and \texttt{HalfCheetah}, respectively.
\item BCQ-L: BCQ-L extends BCQ~\citep{fujimoto2019off} with Lagrangian relaxation to enforce constraints. It is a naive combination of offline RL and safe RL.
\item 
CPQ~\citep{xu2022constraints}: CPQ is the state-of-the-art algorithm specific to offline safe RL. It addresses a stationary constraint via conservative Q-Learning, instead of conditional sequence generation.
\end{itemize}
The hyper-parameter list is placed in \cref{appendixCode} \cref{tab:hyper}.


\subsection{Baseline Comparisons}

In this experiment, the broad applicability of our method is highlighted. Empirically, SaFomer is competitive with state-of-the-art baselines in the discounted setting, and still performs well in the undiscounted setting where all the baselines fail to satisfy the constraints, as shown in \cref{tab:baselinecompare}.

We first evaluate their performance under the discounted setting with a uniform threshold $d=400$ across the different Halfcheetah datasets.
The results show that the proposed SaFormer adheres to the cost limit more strictly than other algorithms and achieves competitive reward returns simultaneously. On the contrary, BCQ-L suffers from oscillation in the learning process due to the changing Lagrangian multipliers and hardly satisfies the constraints. Despite one case of constraint violation, CPQ outperforms SaFormer in terms of cumulative rewards due to the explicit policy optimization based on Q functions. 

We then perform a comparative evaluation under the undiscounted setting and apply a uniform threshold $d=4300$ across different tasks. Note that, we change the cost limit to $4490$ in the HalfCheeta\_medium dataset since the original value is completely out of the offline data distribution, and we will present the experiments in this circumstance later. The results show that only the proposed SaFormer still adheres to the constraints. By contrast, BCQ-L and CPQ are extremely unstable and yield infeasible policies when $\gamma_c = 1$, which is the inherent issue of Q-Learning. 

It is worth mentioning that DT converges to a sub-optimal solution  on the HalfCheetah\_medium\_expert dataset, while SaFormer can avoid the tramp and achieve significantly better performance with a lower cost return instead. To some extent, it reveals that SaFormer is not a simple behaviour cloning conditioned by cost but searches for optimal actions in the constrained space.

\begin{table*}[t]
\begin{center}
\begin{small}
\begin{sc}
\renewcommand\arraystretch{1.15}
\resizebox{1\textwidth}{22mm}{
\begin{tabular}{|c|c|cc|cc|cc|}
\hline
 \multicolumn{2}{|c|}{\multirow{2}*{Dataset}}  &\multicolumn{2}{c|}{20\%Percentile} & \multicolumn{2}{c|}{30\%Percentile} & \multicolumn{2}{c|}{50\%Percentile}   \\
  \cline{3-8}
  \multicolumn{2}{|c|}{~}  & $\zeta_\phi$ ~(\Checkmark) & $\zeta_\phi$ ~(\XSolidBrush)&  $\zeta_\phi$ ~(\Checkmark) & $\zeta_\phi$ ~(\XSolidBrush) & $\zeta_\phi$ ~(\Checkmark) & $\zeta_\phi$ ~(\XSolidBrush)  \\
\hline

\multirow{3}*{\makecell[c]{halfcheetah \\ medium\_expert}} 
& reward & $9121.95\pm 397.5$& $11441.5\pm 12.35$& $9161.5\pm 385.65$&  $11445.21\pm 27.3$& $9296.29 \pm 249.52$ & $11460.68\pm 21.44$ \\
~ & cost & \cellcolor{mygreen!20}$4203.95\pm 2.36$& \cellcolor{myred!20}$4304.72\pm 2.18$&\cellcolor{mygreen!20} $4211.91\pm 9.08$& \cellcolor{myred!20}$4308.48\pm 1.58$& \cellcolor{mygreen!20}$4229.4 \pm 17.35$ & \cellcolor{myred!20} $4305.78\pm 2.87$ \\
~ & limit   & $4215$ & $4215$ & $4224$  & $4224$   & $4266$    & $4266$  \\
\hline
\multirow{3}*{\makecell[c]{walker2d \\ medium\_expert}} 
& return   & $3076.96\pm 101.41$& $3597.44\pm 179.95$& $4278.38\pm 302.14$& $3913.53\pm 302.42$& $4479.84\pm 323.36$& $3989.87\pm 145.14$   \\
~ & cost   & \cellcolor{mygreen!20}$2989.44\pm 75.12$& \cellcolor{myred!20}$3518.32\pm 159.18$& \cellcolor{mygreen!20}$3567.39\pm 174.72$&\cellcolor{mygreen!20} $3576.96\pm 214.27$& \cellcolor{mygreen!20}$3665.75\pm 231.09$& \cellcolor{mygreen!20}$3667.21\pm 90.22$     \\
~ & limit  & $3154$ & $3154$   & $3745$& $3745$ & $3778$    & $3778$ \\
\hline

\multirow{3}*{\makecell[c]{hopper \\ medium\_expert}} 
& reward  &  $1191.74\pm 30.2$& $1671.21\pm 41.19$& $1312.42\pm 25.91$& $1663.75\pm 30.31$& $1467.81\pm 41.64$  & $1677.25\pm 50.03$  \\
~ & cost  & \cellcolor{mygreen!20}$639.62\pm 15.19$&  \cellcolor{myred!20}$756.71\pm 18.93$& \cellcolor{mygreen!20}$696.03\pm 14.93$&  \cellcolor{myred!20}$761.01\pm 12.35$& \cellcolor{mygreen!20}$732.89\pm 21.21$& \cellcolor{mygreen!20}$777.59\pm 25.22$ \\
~ & limit & $683$ & $683$   & $742$ & $742$ & $856$    & $856$   \\
\hline

\end{tabular}
}
\caption{Ablation study on $\zeta_\phi$. We report the performance of SaFormer with and without the posterior safety verification at three different thresholds (20\%,30\%,50\% percentiles, respectively). A full version of the results on all datasets can be found in \cref{appendixT2} \cref{tab:ablation-online}.}
\label{tab:ablation-online-main}
\end{sc}
\end{small}
\end{center}
\vspace{-0.2cm}
\end{table*}

\subsection{Offline Constraint Adaptation}

In this experiment, we demonstrate that SaFormer is robust and flexible against varying in-range constraint thresholds of the offline data without retraining, which is of great attraction to real-world applications.

We train SaFormer on the fixed dataset and use the $10\%,20\%,30\%,50\%$ percentiles of the cost return of the offline trajectories as contextual tokens to evaluate its applicability toward different safety requirements. \cref{tab:zero-shot-main} reports its constraint adaptation performance when pre-trained on HacfCheetah datasets. SaFormer satisfies 11 of 12 cost limits in the experiment, which holds a $91.6\%$ constraint satisfaction rate. In general, SaFormer can identify different cost limits and is able to yield feasible solutions accordingly. 

The only instance of failure in all 12 cases is when using the 10\% percentile of HalfCheetah\_medium\_replay dataset as the threshold ($d = 3928$). Similar situations also occur in the other two environments. A possible reason is that the samples in the replay dataset ($202$ rollouts) are very limited compared with the medium and expert datasets ($> 1000$ rollouts). Thus, the constrained action space might be under-explored for our Transformer-based architecture. 

\subsection{Online Constraint Fine-tuning}

In this experiment, we apply \cref{alg:saformer3} to deal with the under-explored constraints in the aforementioned dataset and show the efficacy of the online fine-tuning technique against those out-of-distribution (OOD) constraints.

At first, we discard samples whose cost returns are less than the previous unmet 10\% percentile ($d=3928$) and regard such thresholds as OOD constraints. Then, we compare the generalizablity of SaFormer via offline pre-training and online fine-tuning. \cref{tab:online-main} shows the mean cost return of the above two types of SaFormer with respect to both in-sample and out-of-sample constraints. We decrease the out-of-sample constraint threshold at equal intervals and witness a drop in the final cost return achieved by offline SaFormer. Nevertheless, the OOD constraints are hardly satisfied via pure offline training. By contrast, the online SaFormer shrinks the expected cost limits to generate new conservative samples and reduces the overall cost below the corresponding  thresholds after fine-tuning. The online sample consumption is acceptable and we insert 200 new trajectories to satisfy all the constraints listed in \cref{tab:online-main}.

\subsection{Ablation Study}


In this experiment, we conduct ablation study to demonstrate the necessity of the posterior safety verification and the sensitivity of the proposal batchsize.

\vspace{-0.2cm}
\paragraph{Posterior safety verification} We first investigate the necessity of the critic $\zeta_\phi$, which evaluates the long-term cost return and filters out unsafe actions. \cref{tab:ablation-online-main} reports the performance of SaFormer at 20\%, 30\% and 50\% percentiles of in-range thresholds with and without the posterior safety verification, respectively. The results show that SaFormer fulfills the hard constraints well when it is equipped with $\zeta_\phi$. On the contrary, the naive actor $\pi_\theta$ will only lead to near constraint-satisfying policies and still holds the $55.6\%$ of the proportion that violates the constraint. This phenomenon is more pronounced when the constraint is tightened. Furthermore, even if the actor can directly yield feasible trajectories under certain constraints, their cumulative rewards are inferior to SaFormer with posterior safety verification. The reason is that SaFormer samples a batch of candidates and executes the one with the highest RTG by leveraging $\zeta_\phi$, which may boost the reward performance while preserving the feasibility.

\begin{figure}
\begin{center}
\centerline{\includegraphics[width=\columnwidth,trim=10 0 0 0,clip]{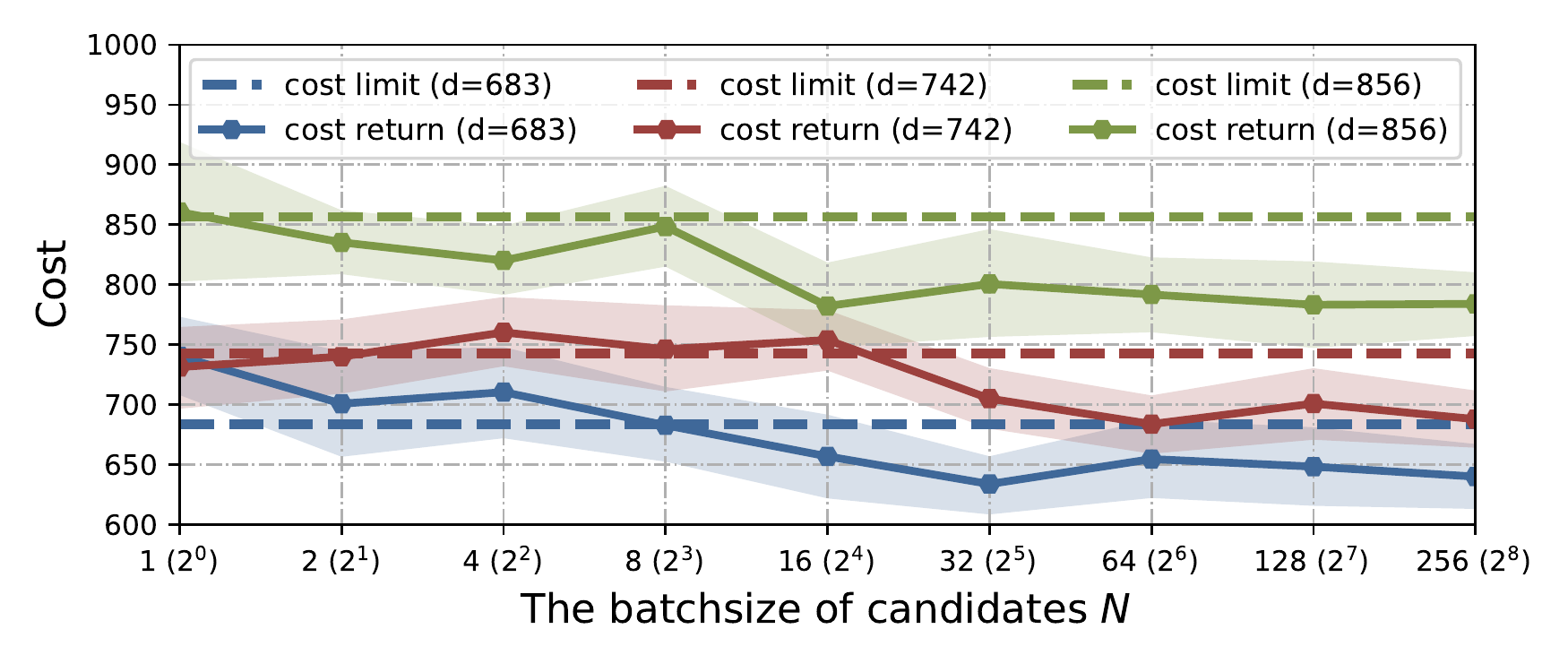}}
\vspace{-0.175cm}
\caption{Sensitivity study on RTG and action proposal batchsize $N$. The curve denotes the eventual cost return with respect to different $N$. The dashed line in the corresponding color denotes the cost limit of the Hopper\_medium\_expert dataset.}
\label{sensitivity}
\end{center}
\vspace{-0.3cm}
\end{figure}

\vspace{-0.2cm}
\paragraph{Proposal batchsize} At last, we study the effect of RTG and action proposal batchsize $N$. We evaluate the constraint satisfaction at 20\%, 30\%, and 50\% percentiles of in-range thresholds ($d=683, 742, 856,$ respectively) on the Hopper\_medium\_expert dataset with respect to different $N$. \cref{sensitivity} confirms that the cost return is steady and constraint-satisfying when the actor proposes a large batch of candidates. By contrast, SaFormer may exceed the cost limit during the execution when the batch size $N$ is relatively small. It is concluded that the mapping from cost to reward is not a one-to-one function and requires a wide range of samples to search for the optimal solution. The eventual returns converge if the batchsize is sufficiently large.

\section{Conclusion}
We present SaFormer in this paper, which, to the best of our knowledge, is the first sequence modeling approach to offline safe RL. Casting constraints as contextual tokens, SaFormer is competitive with state-of-the art algorithms in terms of reward performance, but more robust and flexible toward varying safety requirements. We believe such the properties is of great practical relevance in real-world problems. As the future work, more complicated offline datasets and tasks are required to better evaluate the proposed approach. Besides, we are also dedicated to overcome the fundamental assumptions in the problem setup to enhance SaFormer's risk-awareness and extend SaFormer to multi-constraint scenarios to reduce its limitations in safety-critical tasks.



\bibliography{example_paper}
\bibliographystyle{icml2022}

\clearpage

\onecolumn
\icmltitle{Appendix for ``SaFormer: A Conditional Sequence Modeling Approach to Offline Safe Reinforcement Learning"}

\appendix

\section{Dataset and task visualizations}
\label{appendixA}
\begin{figure}[H]
      \centering
    \subcaptionbox{Hopper\_medium}
        {\includegraphics[width=0.3\linewidth,trim=0 0 0 0,clip]{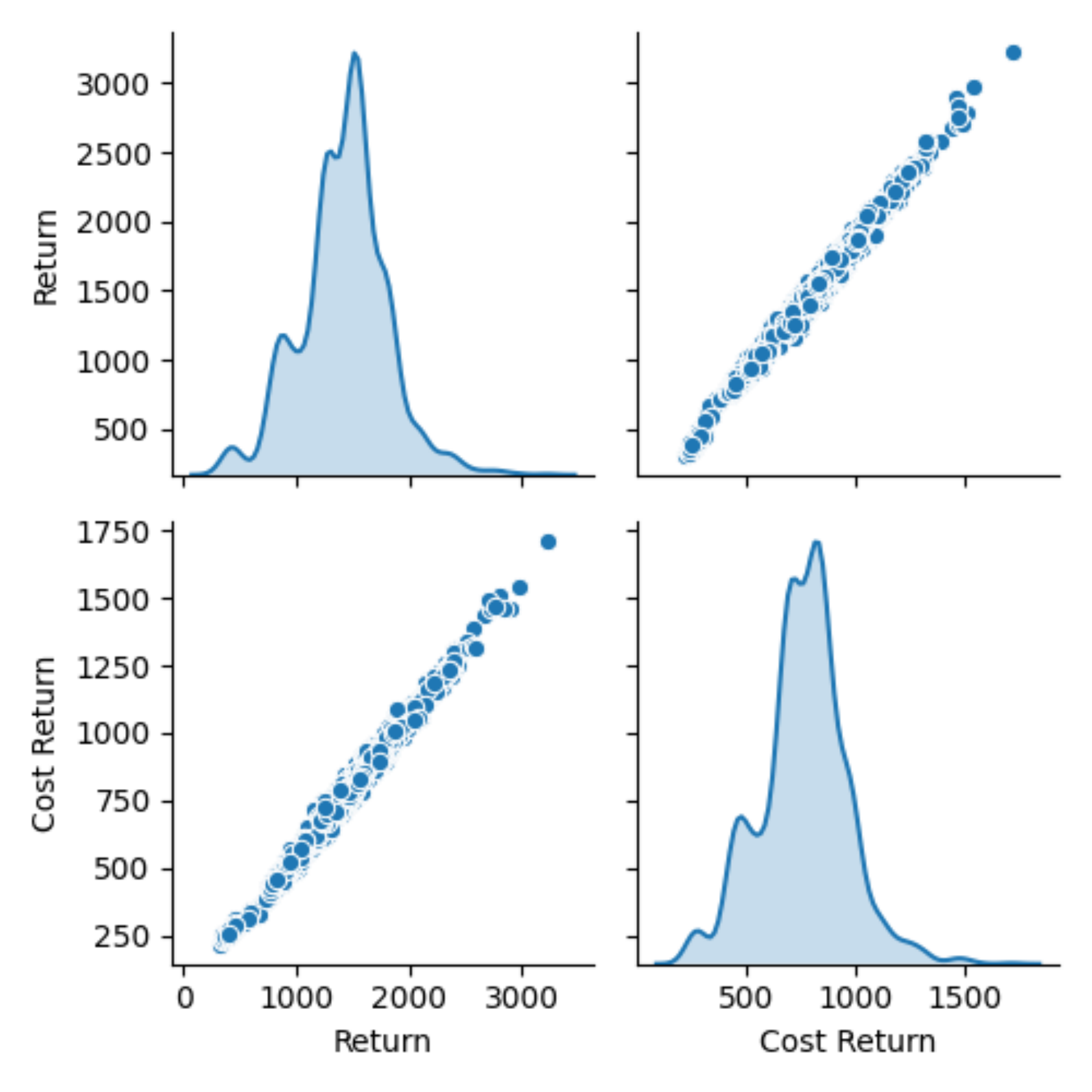}}
     \subcaptionbox{Hopper\_medium\_replay}
        {\includegraphics[width=0.3\linewidth,trim=0 0 0 0,clip]{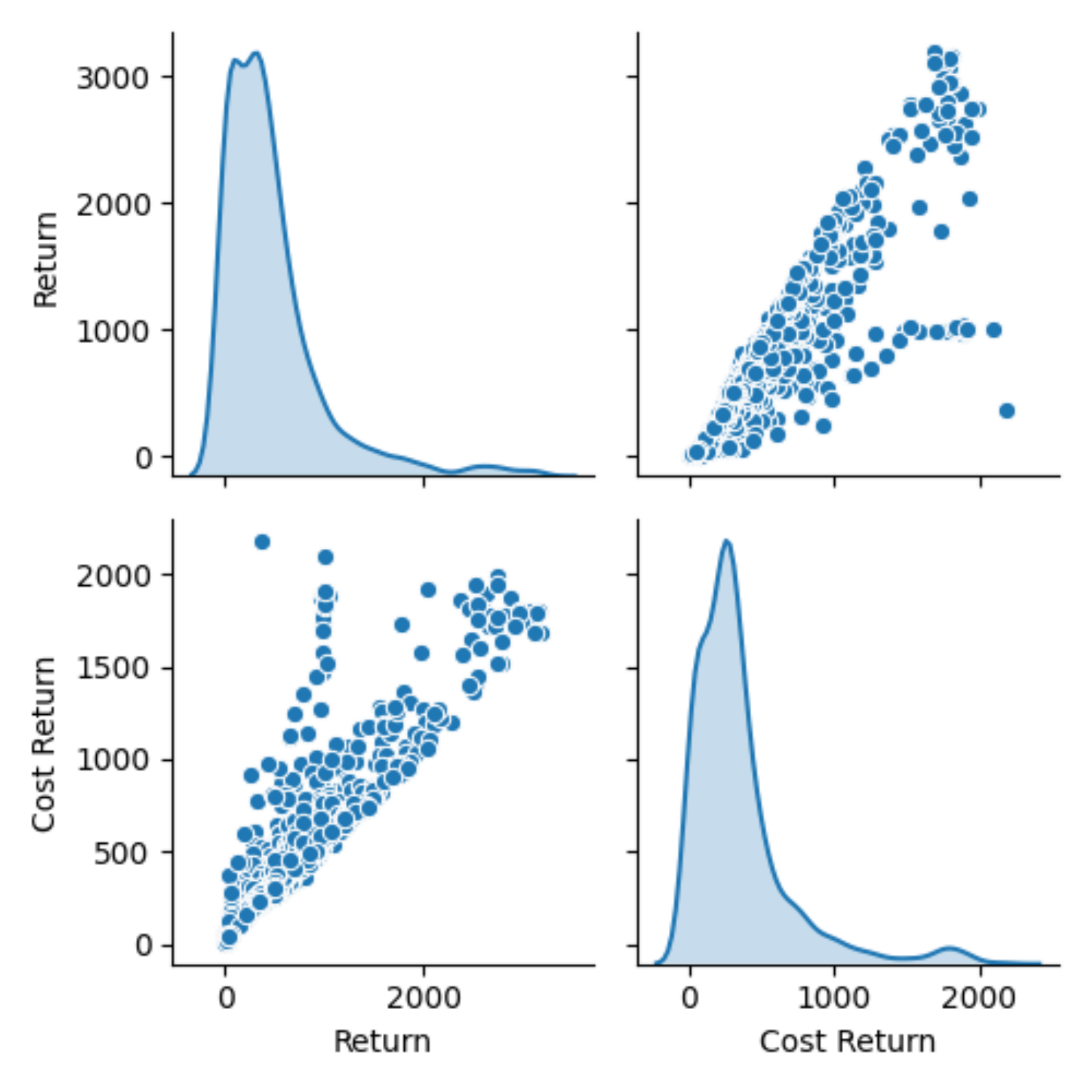}}
        \subcaptionbox{Hopper\_medium\_expert}
        {\includegraphics[width=0.3\linewidth,trim=0 0 0 0,clip]{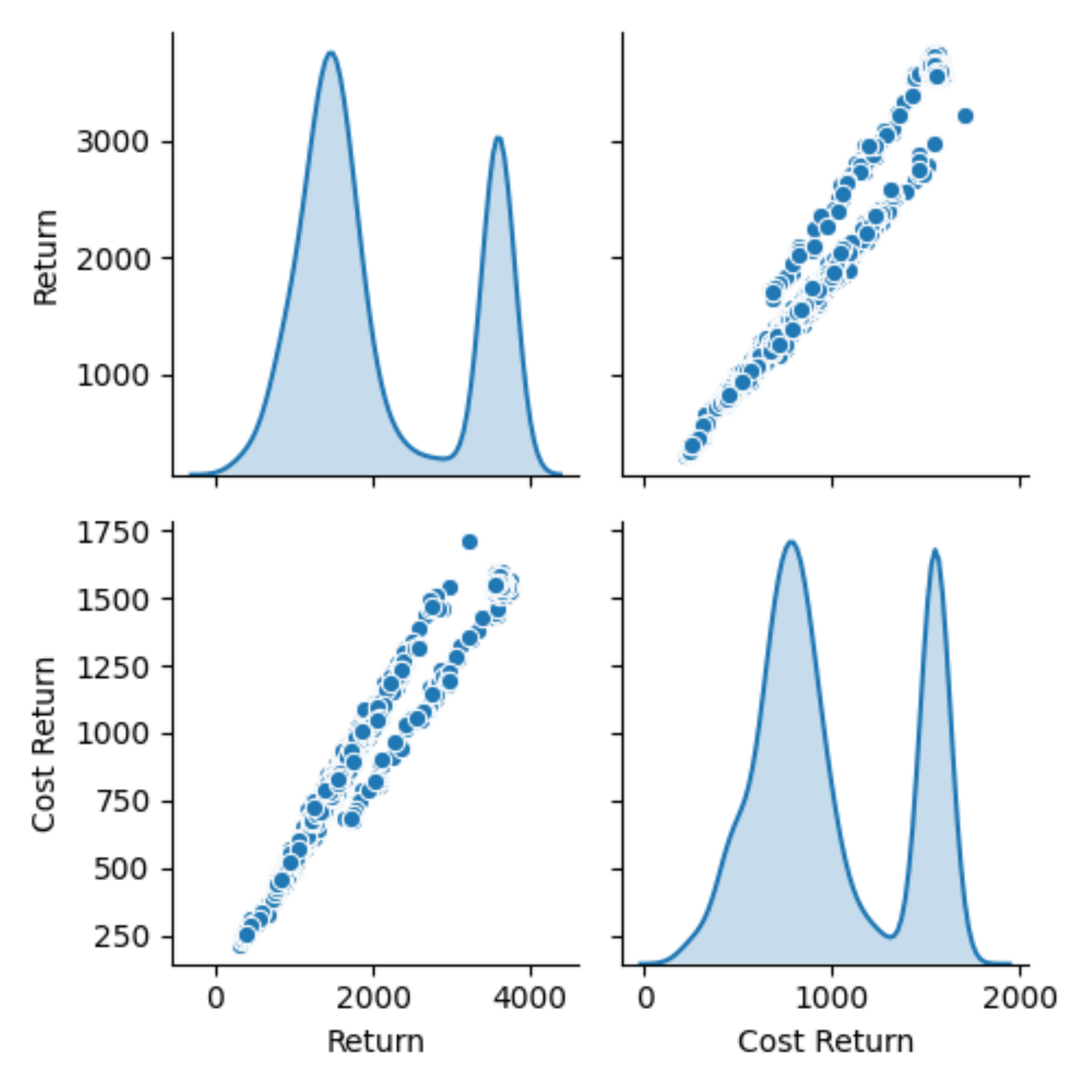}}
    \subcaptionbox{Walker2d\_medium}
        {\includegraphics[width=0.3\linewidth,trim=0 0 0 0,clip]{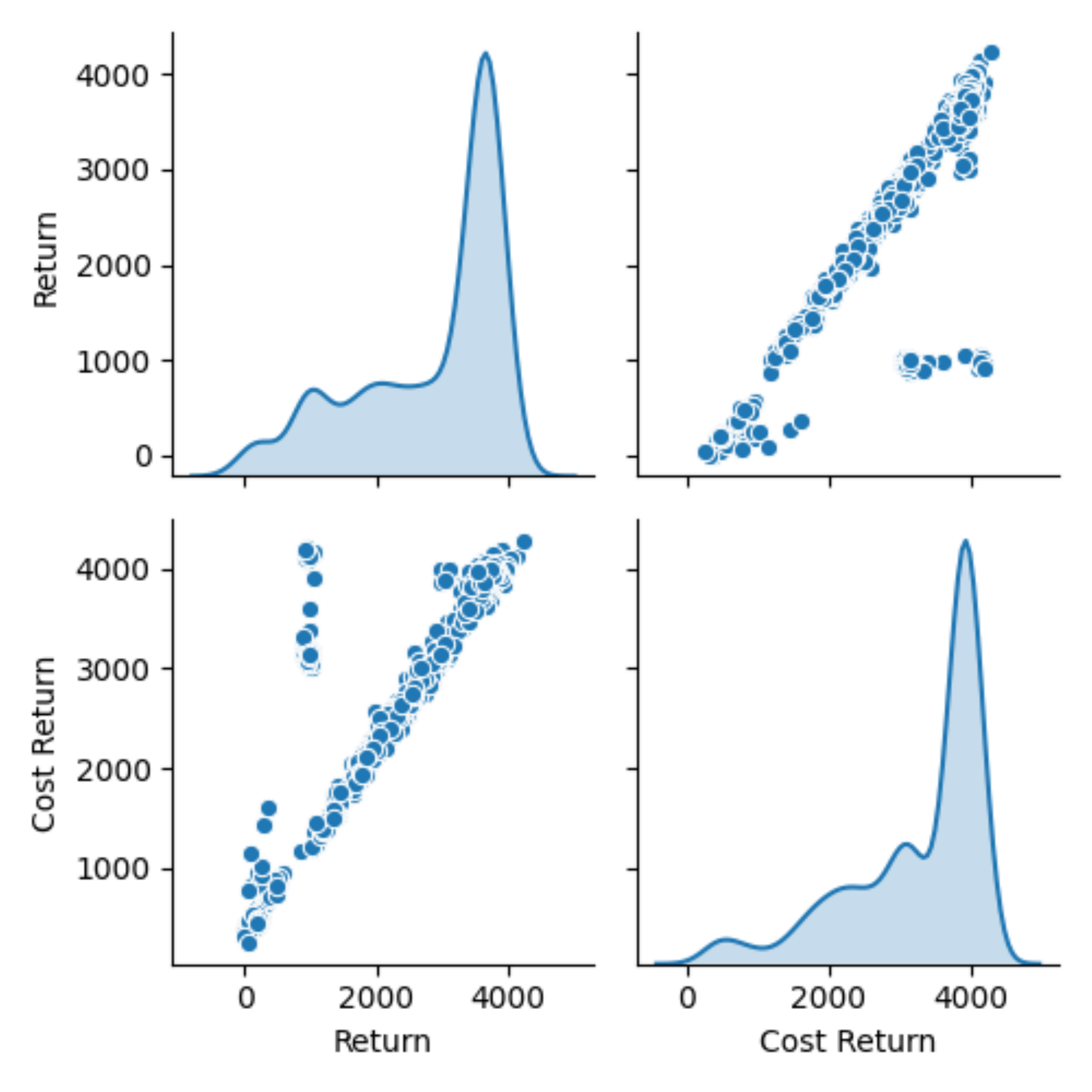}}
     \subcaptionbox{Walker2d\_medium\_replay}
        {\includegraphics[width=0.3\linewidth,trim=0 0 0 0,clip]{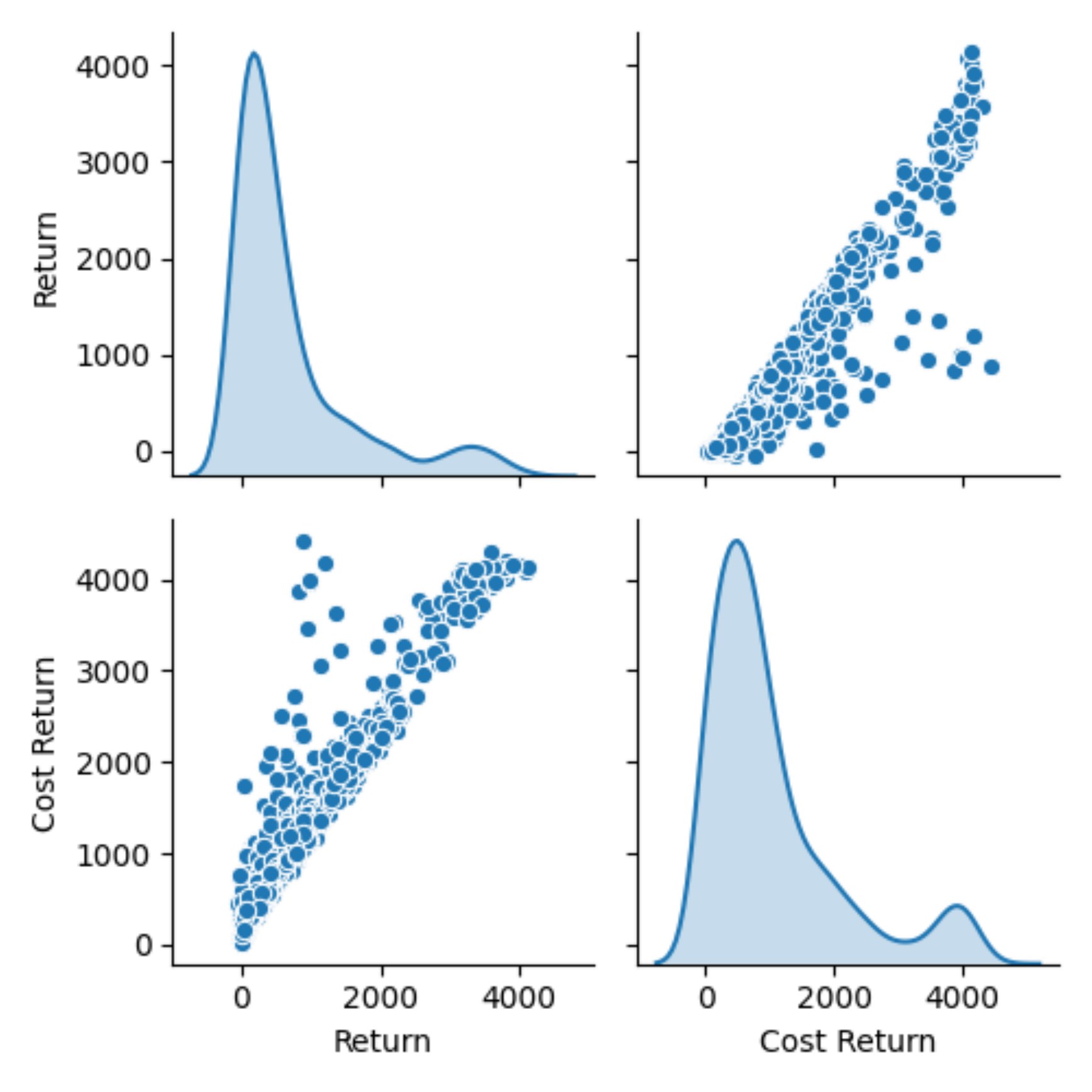}}
        \subcaptionbox{Walker2d\_medium\_expert}
        {\includegraphics[width=0.3\linewidth,trim=0 0 0 0,clip]{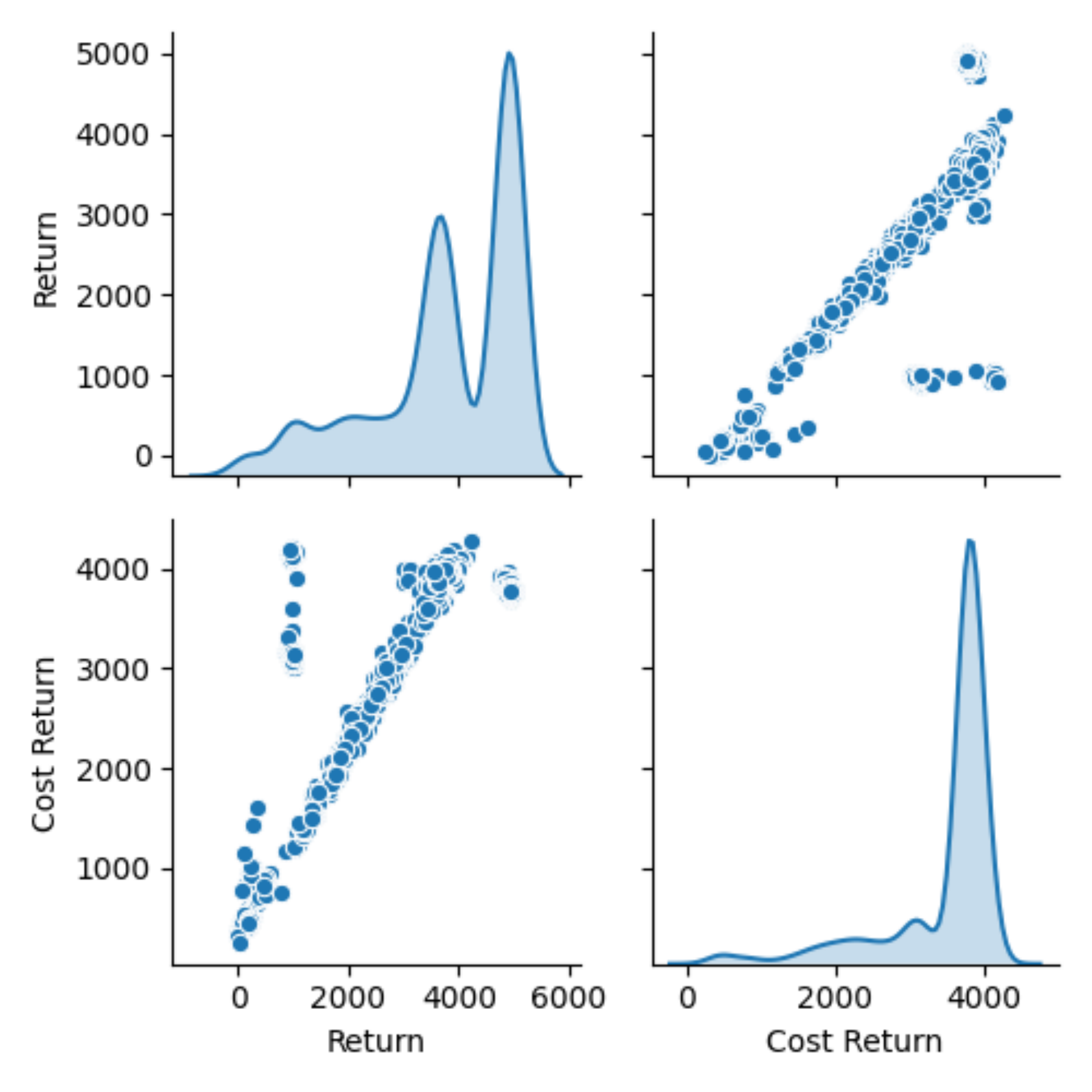}}
    \subcaptionbox{HalfCheetah\_medium}
        {\includegraphics[width=0.3\linewidth,trim=0 0 0 0,clip]{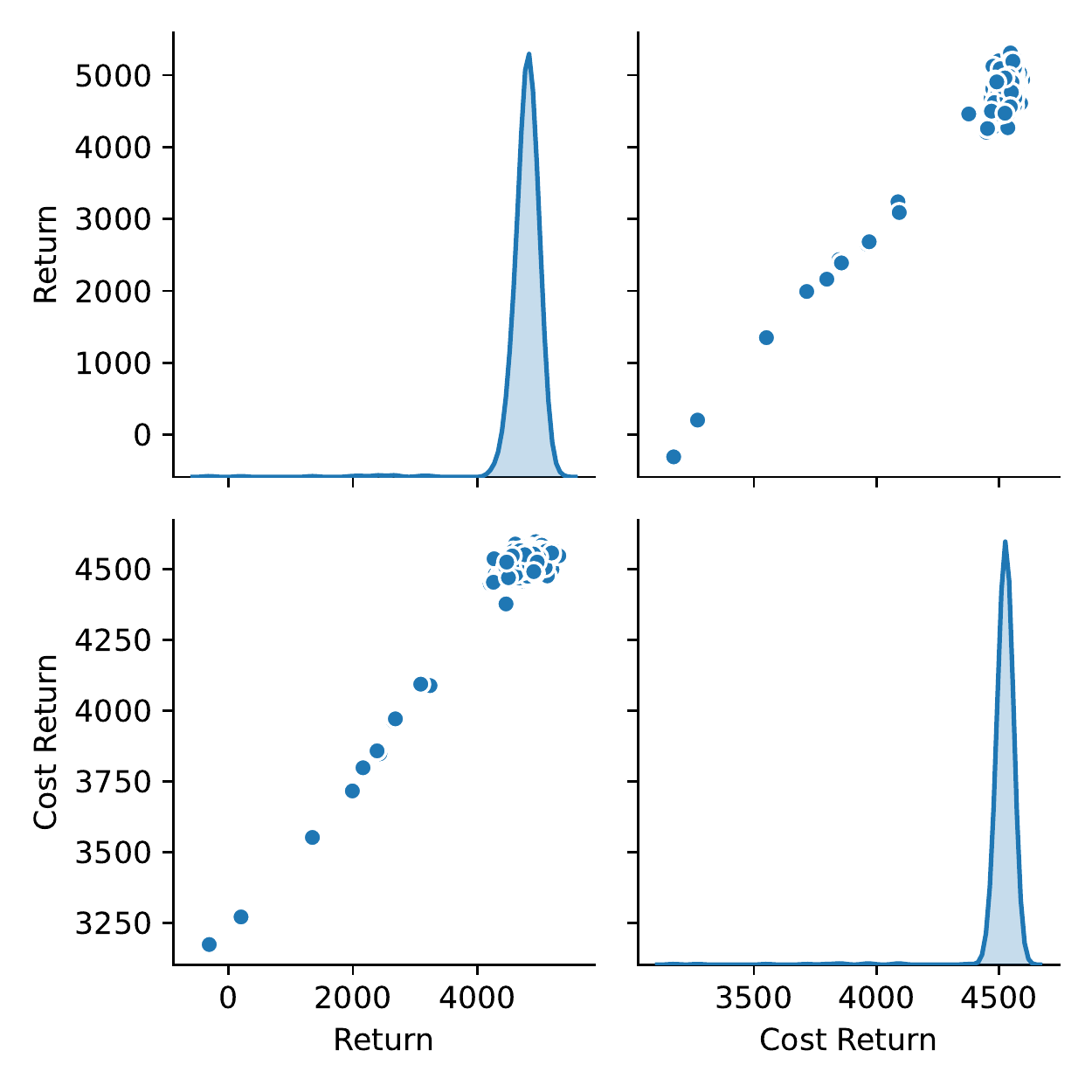}}
     \subcaptionbox{HalfCheetah\_medium\_replay}
        {\includegraphics[width=0.3\linewidth,trim=0 0 0 0,clip]{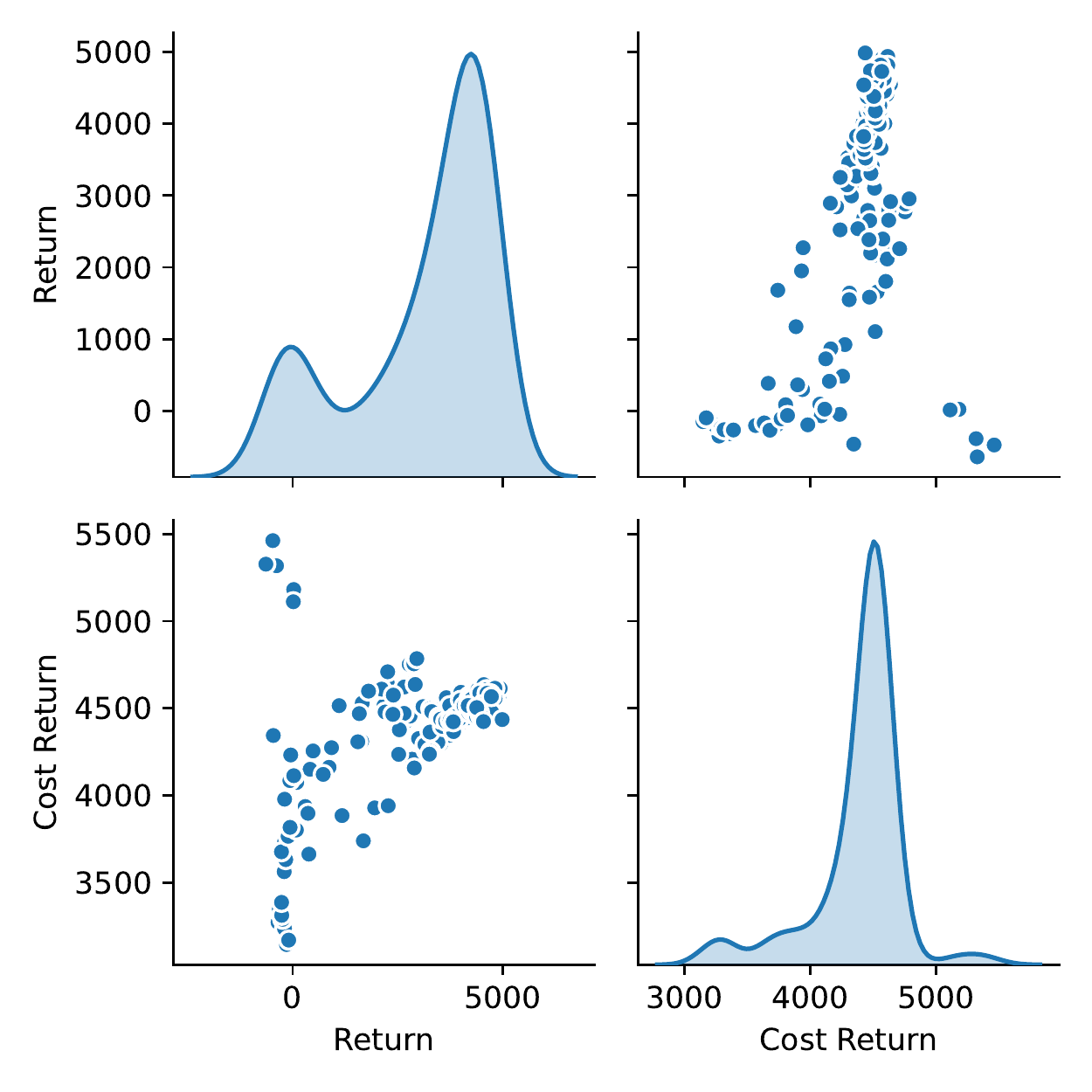}}
        \subcaptionbox{HalfCheetah\_medium\_expert}
        {\includegraphics[width=0.3\linewidth,trim=0 0 0 0,clip]{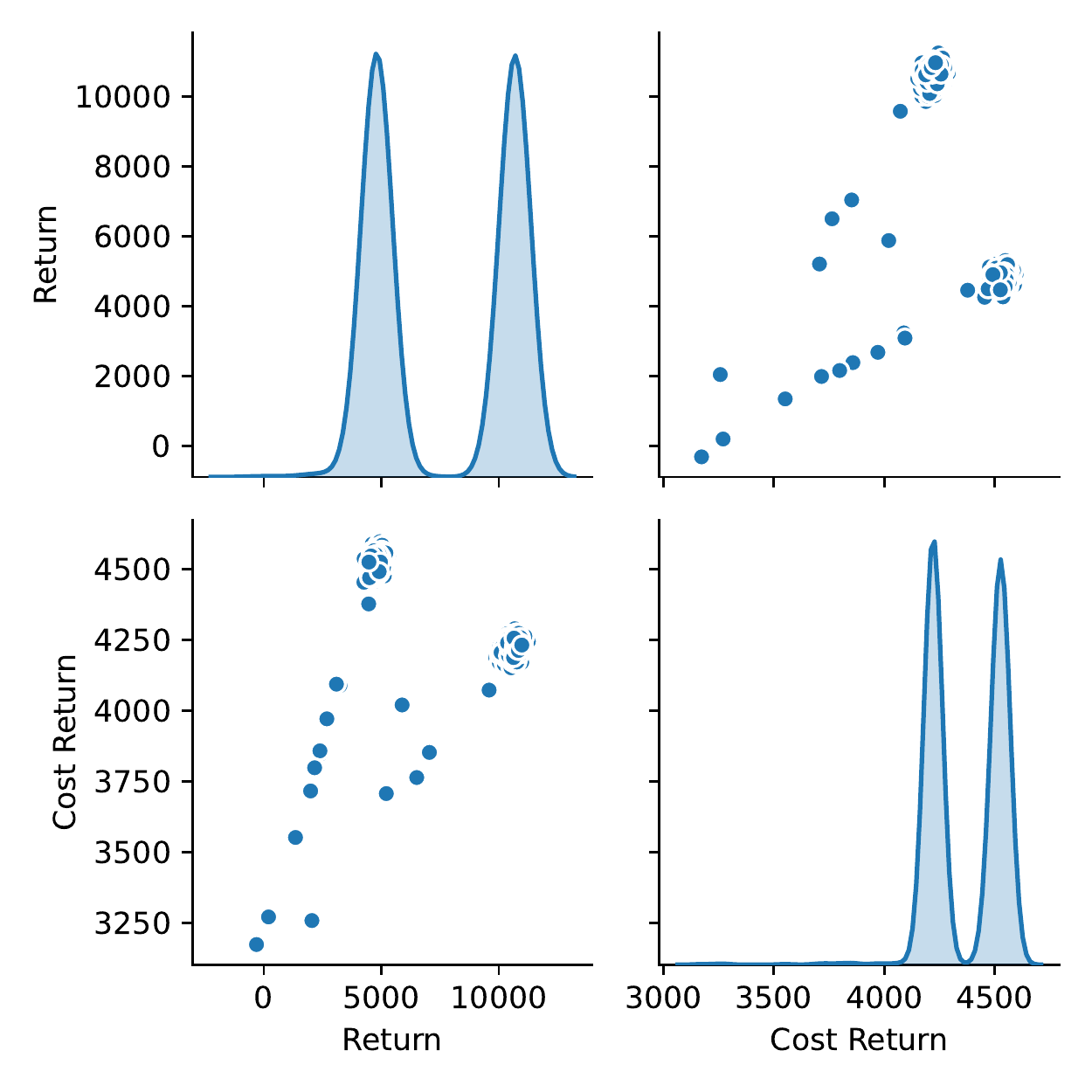}}
      \caption{Pair-plots of reward and cost returns on the D4RL datasets.}
      \label{fig:distribution}   
\end{figure}

\begin{table*}[t]
\begin{center}
\begin{small}
\begin{sc}
\renewcommand\arraystretch{1.15}
\resizebox{1\textwidth}{22.5mm}{
\begin{tabular}{|c|cccccc|}
\toprule
 \small{Dataset} & \small{Min} & \small{10\%Percentile} & \small{20\%Percentile} & \small{30\%Percentile}  & \small{50\%Percentile} & \small{Max}   \\
\midrule
\makecell[c]{halfcheetah\_medium}
 & $3172$  & $4490$ & $4503$ & $4511$ & $4526$ &$4600$ \\

\makecell[c]{halfcheetah\_medium\_replay}
 & $3145$ & $3928$ & $4257$ & $4422$ & $4493$ &$5461$\\

\makecell[c]{halfcheetah\_medium\_expert}
& 3172 & $4201$& $4215$ & $4224$ & $4266$  &4600\\

\makecell[c]{walker2d\_medium}
 & $246$ & $1877$ & $2450$ & $3043$ & $3839$  &$4283$ \\

\makecell[c]{walker2d\_medium\_replay}
&$8$  &  $75$ & $251$ & $432$ & $705$ & 4432\\

\makecell[c]{walker2d\_medium\_expert}
 &$246$  & $2343$ & $3154$ & $3745$ & $3778$  & $4283$ \\

\makecell[c]{hopper\_medium}
&$218$ & $483$ & $618$ & $685$ & $772$ & $1710$ \\

\makecell[c]{hopper\_medium\_replay}
&2  & $35$ & $90$ & $184$ & $268$  &$2186$\\

\makecell[c]{hopper\_medium\_expert}
 &$218$ & $546$ & $683$ & $742$ & $856$ & $1710$  \\
\bottomrule
\end{tabular}
}
\caption{Threshold settings (10\%,20\%,30\%,50\% percentiles) on the D4RL datasets.}
\label{tab:threshold}
\end{sc}
\end{small}
\end{center}
\vspace{-0.25cm}
\end{table*}

\section{Hyper-parameters}
\label{appendixCode}

The hyper-parameters are listed below.

\begin{table}[H]
\begin{center}
\begin{small}
\begin{sc}
\renewcommand\arraystretch{1.15}
\resizebox{1\textwidth}{40.5mm}{
\begin{tabular}{|c|p{2cm}<{\centering}p{2cm}<{\centering}p{2cm}<{\centering}p{2cm}<{\centering}|}
\toprule
 Hyper-parameters & SaFormer & DT & BCQ-L & CPQ   \\
\midrule
Subsequence length $K$
 & 20  & 20 & / & / \\
Number of attention blocks 
 & 3  & 3 & / & / \\
 Embedding dimension  & 128  & 128 & / & /\\
 Hidden layer  & 128  & 128 & (256,256) & (256,256) \\
 Dropout ratio  & 0.1  & 0.1 & 0.1 & 0.1 \\
 Training batchsize & 128  & 128 & 128 & 128\\
 Learning rate & 10E-4& 10E-4& 10E-4& 10E-4\\
 Learning rate Decay & 10E-4& 10E-4& 10E-4& 10E-4\\
 Proposal batchsize $N$ & 128  & / & /& /  \\
 Penalty factor $\lambda$ & 0.25 & / & /& / \\
 Attenuation factor $\alpha$ & 0.95 & / & /& / \\
 Initial Lagrangian multiplier  & / & /  & 0.1 & /\\
 Lagrangian multiplier Learning rate & / & / & 3E-4 & /\\
 Polyak Averaging factor $\tau$ & / & / & 0.05 & 0.05\\
 OOD penalty $\alpha$~\cite{xu2022constraints} & / & / & / & 5.0\\
 VAE penalty $\beta$~\cite{xu2022constraints} & / & / & / & 1.5\\

\bottomrule
\end{tabular}
}
\caption{Hyper-parameter lists for baseline comparisons.}
\label{tab:hyper}
\end{sc}
\end{small}
\end{center}
\vspace{-0.25cm}
\end{table}

\clearpage
\section{Empirical details}
\label{appendixT2}
\begin{table}[H]
\begin{center}
\begin{small}
\begin{sc}
\renewcommand\arraystretch{1.15}
\resizebox{1\textwidth}{50mm}{
\begin{tabular}{|c|c|cccc|c|}
\hline
 \multicolumn{2}{|c|}{\small{Dataset}}  & \small{10\%Percentile} & \small{20\%Percentile} & \small{30\%Percentile}  & \small{50\%Percentile}  & DT(Reference)   \\
\hline

\multirow{3}*{\makecell[c]{halfcheetah \\ medium}} 
& reward    & $4661.2\pm 52.46$& $4713.52\pm 26.83$& $4718.44\pm 22.09$& $4724.89\pm 27.18$  & $4986.64\pm 5.56$ \\
~ & cost    &$4487.2\pm 6.05$& $4498.22\pm 2.57$& $4500.52\pm 1.65$& $4508.83\pm 2.79$  & $4531.89\pm 1.45$ \\
~ &  limit  & $4490$ & $4503$ & $4511$ & $4526$ & --\\

\hline
\multirow{3}*{\makecell[c]{halfcheetah \\ medium\_replay}} 
& reward    & $2202.69\pm 148.39$& $2208.41\pm 90.14$ & $3363.11\pm 165.89$& $3712.08\pm 104.74$   & $4270.07 \pm 16.11$ \\
~ & cost    & $4203.0\pm 17.46$&$4234.15\pm 13.36$ & $4381.42\pm 16.94$&$4411.84\pm 9.29$ & $4448.84 \pm 6.8$\\
~ & limit   & $3928$ & $4257$ & $4422$ & $4493$ & --\\

\hline

\multirow{3}*{\makecell[c]{halfcheetah \\ medium\_expert}} 
& reward    & $9016.82\pm 124.45$& $9121.95\pm 397.5$& $9161.5\pm 385.65$& $9296.29\pm 249.52$ & $5053.89 \pm 30.17$\\
~ & cost    &$4181.25\pm 16.63$& $4203.95\pm 2.36$& $4211.91\pm 9.08$& $4229.4\pm 17.35$ &   $4478.89 \pm 3.52$ \\
~ & limit   & $4201$& $4215$ & $4224$ & $4266$ &  -- \\

\hline
\multirow{3}*{\makecell[c]{walker2d \\ medium}} 
& reward       & $2654.84\pm 210.72$& $2752.82\pm 161.99$& $2896.36\pm 72.88$& $3016.77\pm 139.35$ & $3325.07\pm 94.88$\\
~ & cost       & $3038.95\pm 222.47$& $3191.76\pm 138.93$&  $3224.36\pm 67.41$& $3383.85\pm 121.79$ & $3360.02\pm 99.78$ \\
~ & limit      & $1877$ & $2450$ & $3043$ & $3839$ &  -- \\

\hline

\multirow{3}*{\makecell[c]{walker2d \\ medium\_replay}} 
& reward    &  $32.54\pm 20.95$& $155.82\pm 36.2$& $238.5\pm 20.42$& $318.8\pm 42.47$ & $2194.73\pm 176.7$\\
~ & cost    &  $124.55\pm 39.66$& $417.45\pm 105.91$& $585.19\pm 28.27$&$682.98\pm 31.45$ & $2259.91\pm 198.37$ \\
~ & limit   &  $75$ & $251$ & $432$ & $705$ & -- \\

\hline

\multirow{3}*{\makecell[c]{walker2d \\ medium\_expert}}
& reward        & $2938.5\pm 140.71$& $3076.96\pm 101.41$& $4278.38\pm 302.14$& $4479.84\pm 323.36$ & $4972.82\pm 1.01$\\
~ & cost        & $3160.32\pm 130.84$& $2989.44\pm 75.12$&$3567.39\pm 174.72$&$3665.75\pm 231.09$ & $3569.08\pm 0.41$ \\
~ & limit       & $2343$ & $3154$ & $3745$ & $3778$ &--\\

\hline
\multirow{3}*{\makecell[c]{hopper\\ medium}} 
& reward   & $862.21\pm 17.59$& $1037.53\pm 21.69$& $1090.57\pm 23.52$& $1245.57\pm 25.04$  & $2018.02\pm 79.05$\\
~ & cost   &$477.39\pm 9.12$& $568.83\pm 12.04$&$599.08\pm 10.59$&$671.1\pm 13.18$  &  $907.24\pm 38.19$  \\
~ & limit  & $483$ & $618$ & $685$ & $772$& --\\

\hline

\multirow{3}*{\makecell[c]{hopper\\ medium\_replay}} 
& reward  & $39.28\pm 6.35$& $71.26\pm 14.0$& $244.93\pm 32.71$& $351.12\pm 59.37$  & $904.97\pm 193.08$\\
~ & cost  & $50.03\pm 3.95$& $89.52\pm 14.47$& $207.42\pm 13.25$& $259.67\pm 11.71$ & $346.6\pm 73.01$\\
~ & limit & $35$ & $90$ & $184$ & $268$ & - \\

\hline
\multirow{3}*{\makecell[c]{hopper\\ medium\_expert}} 
& reward    & $986.29\pm 29.72$& $1191.74\pm 30.2$& $1312.42\pm 25.91$& $1467.81\pm 41.64$  & $3274.14\pm 176.99$ \\
~ & cost    &$535.13\pm 18.88$&$639.62\pm 15.19$&$696.03\pm 14.93$&$732.89\pm 21.21$ & $1113.92\pm 65.25$\\
~ & limit   & $546$ & $683$ & $742$ & $856$ & -- \\

\hline
\end{tabular}
}
\caption{Full experiment results of constraint adaptation.}
\label{tab:zero-shot}
\end{sc}
\end{small}
\end{center}
\vspace{-0.25cm}
\end{table}

\begin{table}[H]
\begin{center}
\begin{small}
\begin{sc}
\renewcommand\arraystretch{1.15}
\resizebox{1\textwidth}{50mm}{
\begin{tabular}{|c|c|cc|cc|cc|}
\hline
 \multicolumn{2}{|c|}{\multirow{2}*{Dataset}}  &\multicolumn{2}{c|}{20\%Percentile} & \multicolumn{2}{c|}{30\%Percentile} & \multicolumn{2}{c|}{50\%Percentile}   \\
  \cline{3-8}
  \multicolumn{2}{|c|}{~}  & $\zeta_\phi$ ~(\Checkmark) & $\zeta_\phi$ ~(\XSolidBrush)&  $\zeta_\phi$ ~(\Checkmark) & $\zeta_\phi$ ~(\XSolidBrush) & $\zeta_\phi$ ~(\Checkmark) & $\zeta_\phi$ ~(\XSolidBrush)  \\
\hline

\multirow{3}*{\makecell[c]{halfcheetah\\medium}} 
& reward     & $4611.15\pm 74.75$& $4884.18\pm 166.87$& $4604.47\pm 110.94$& $4883.92\pm 170.97$& $4724.89 \pm 27.18$  & $4968.35\pm 167.89$ \\
~ & cost    &$4478.43\pm 12.5$& $4590.93\pm 75.71$& $4477.93\pm 20.45$& $4580.49\pm 93.37$&$4508.83 \pm 2.79$ &  $4614.53\pm 74.26$  \\
~ &  limit  & $4503$ & $4503$ & $4511$ & $4511$  &$4526$  &$4526$   \\
\hline
\multirow{3}*{\makecell[c]{halfcheetah\\medium\_replay}}
& reward    & $2208.41\pm 90.14$& $3807.5\pm 72.0$& $3363.11\pm 165.89$& $3646.36\pm 42.99$& $3712.08 \pm 104.74$ & $3470.89\pm 185.94$\\
~ & cost    &$4234.15\pm 13.36$& $4578.56\pm 6.62$& $4381.42\pm 16.94$& $4596.31\pm 11.27$&$4411.84 \pm 9.29$ &  $4579.14\pm 57.62$  \\
~ & limit    &  $ 4257$  & $4257$  & $4422$  & $4422 $ & $4493$   & $4493$  \\
\hline
\multirow{3}*{\makecell[c]{halfcheetah\\medium\_expert}} 
& reward & $9121.95\pm 397.5$& $11441.5\pm 12.35$& $9161.5\pm 385.65$&  $11445.21\pm 27.3$& $9296.29 \pm 249.52$ & $11460.68 \pm 21.44$ \\
~ & cost &$4203.95\pm 2.36$& $4304.72\pm 2.18$& $4211.91\pm 9.08$& $4308.48\pm 1.58$&$4229.4 \pm 17.35$ &  $4305.78\pm 1 2.87$ \\
~ & limit   & $4215$ & $4215$ & $4224$  & $4224$   & $4266$    & $4266$  \\
\hline

\multirow{3}*{\makecell[c]{walker2d\\medium}} 
& reward    & $2752.82\pm 161.99$& $3304.39\pm 150.29$& $2896.36\pm 72.88$& $3416.02\pm 248.65$& $3016.77\pm 139.35$& $3367.41\pm 159.17$ \\
~ & cost    &  $3191.76\pm 138.93$&  $3376.99\pm 143.7$&  $3224.36\pm 67.41$&  $3508.21\pm 240.09$&$3383.85\pm 121.79$&$3439.88\pm 163.23$      \\
~ & limit     & $2450$ & $2450$ & $3043$  & $3043$  & $3839$    & $3839$   \\
\hline
\multirow{3}*{\makecell[c]{walker2d\\medium\_replay}} 
& reward      & $155.82\pm  36.2$& $192.46\pm 14.19$ & $238.5\pm  20.42$& $314.85\pm 51.67$& $318.8\pm  42.47$& $535.65\pm 64.69$ \\
~ & cost    &  $417.45\pm  105.91$&  $482.54\pm 27.56$&  $585.19\pm  28.27$&  $618.22\pm 88.42$& $682.98\pm  31.45$&  $873.69\pm 174.75$  \\
~ & limit     & $251$ & $251$   & $432$ & $432$ & $705$    & $705$  \\

\hline

\multirow{3}*{\makecell[c]{walker2d\\medium\_expert}} 
& reward   & $3076.96\pm 101.41$& $ 3597.44 \pm 179.95$& $4278.38\pm 302.14$& $3913.53 \pm 302.42$& $4479.84\pm 323.36$& $ 3989.87 \pm 145.14$   \\
~ & cost   &$2989.44\pm 75.12$&  $3518.32 \pm 159.18$& $3567.39\pm 174.72$&$3576.96 \pm 214.27$&$3665.75\pm 231.09$& $ 3667.21 \pm 90.22$     \\
~ & limit  & $3154$ & $3154$   & $3745$& $3745$ & $3778$    & $3778$ \\
\hline

\multirow{3}*{\makecell[c]{hopper\\medium}} 
& reward   & $1037.53\pm 21.69$& $1504.02\pm 31.34$& $1090.57\pm 23.52$& $1584.71\pm 60.46$& $1245.57\pm 25.04$& $1659.57\pm 65.07$    \\
~ & cost   &$568.83\pm 12.04$&  $682.61\pm 17.72$&$599.08\pm 10.59$&  $718.98\pm 28.18$& $671.1\pm 13.18$& $757.86\pm 31.93$    \\
~ & limit    & $618$ & $618$&  $685$ & $685$& $772$    & $772$  \\
\hline
\multirow{3}*{\makecell[c]{hopper\\medium\_replay}} 
& reward  & $71.26\pm  14.0$& $244.24\pm 36.01$& $244.93\pm  32.71$& $340.14\pm 21.47$& $351.12\pm  59.37$& $514.83\pm 108.93$   \\
~ & cost  &  $89.52\pm  14.47$&  $241.02\pm 25.91$&  $207.42\pm  13.25$&  $281.49\pm 12.8$& $259.67\pm  11.71 $&  $290.89\pm 46.99$   \\
~ & limit  & $90$ & $90$  & $184$ & $184$ & $268$    & $268$   \\
 \hline
 \multirow{3}*{\makecell[c]{hopper\\medium\_expert}} 
& reward  &  $1191.74\pm 30.2$& $1671.21 \pm 41.19$& $1312.42\pm 25.91$& $1663.75\pm 30.31$& $1467.81\pm 41.64$& $1677.25\pm 50.03$    \\
~ & cost  &$639.62\pm 15.19$&  $756.71 \pm  18.93$&$696.03\pm 14.93$&  $761.01\pm 12.35$& $732.89\pm 21.21$& $777.59\pm 25.22$ \\
~ & limit & $683$ & $683$   & $742$ & $742$ & $856$    & $856$   \\
\hline
\end{tabular}
}
\caption{Full experiment results of ablation study.}
\label{tab:ablation-online}
\end{sc}
\end{small}
\end{center}
\vspace{-0.25cm}
\end{table}


\end{document}